\documentclass[10pt,twocolumn,letterpaper]{article}

\usepackage[pagenumbers]{cvpr}      
\usepackage{pifont}
\usepackage{graphicx}
\usepackage{amsmath}
\usepackage{amssymb}
\usepackage{array}
\usepackage{multirow, nicefrac}
\usepackage{cite}
\usepackage{color, xcolor}
\usepackage{makecell}
\usepackage{multicol}
\usepackage{bbm}
\usepackage{xcolor}
\usepackage{xspace}
\usepackage{float}

\usepackage{enumitem}
\usepackage{colortbl}

\definecolor{cvprblue}{rgb}{0.21,0.49,0.74}
\usepackage[pagebackref,breaklinks,colorlinks,citecolor=cvprblue]{hyperref}
\definecolor{mygray}{gray}{.92}

\newlength\savedwidth

\newlength\savewidth
\newcommand\shline{\noalign{\global\savewidth\arrayrulewidth
                           \global\arrayrulewidth 0.5pt}%
                  \hline
                  \noalign{\global\arrayrulewidth\savewidth}}

\newcommand{\zt}[1]{z_{#1}}
\newcommand{\model}{\epsilon_\theta}
\newcommand{\conditioner}{\tau_\theta}
\newcommand{\encoder}{\mathcal{E}}
\newcommand{\expec}{\mathbb{E}}

\def\paperID{7379} 
\def\confName{CVPR}
\def\confYear{2024}

\title{Paragraph-to-Image Generation with 
Information-Enriched 
Diffusion Model}

\author{
Weijia Wu\textsuperscript{$1,$}\textsuperscript{$2,$}\textsuperscript{$3$}
  \quad Zhuang Li\textsuperscript{$1$}
  \quad Yefei He\textsuperscript{$2$} 
  \quad Mike Zheng Shou\textsuperscript{$3$}$^\dagger$
  \quad Chunhua Shen\textsuperscript{$2$}
  \quad Lele Cheng\textsuperscript{$1$}
  \\
  \quad Yan 
  Li\textsuperscript{$1$}
  \quad Tingting Gao\textsuperscript{$1$}
  \quad Di Zhang\textsuperscript{$1$}
  \\[0.2cm]
  \textsuperscript{1}Kuaishou Technology
  ~~~
  \textsuperscript{2}Zhejiang University 
  ~~~
  \textsuperscript{3}Show Lab, National University of Singapore
}

\begin{document}

\twocolumn[{
\renewcommand\twocolumn[1][]{#1}
\maketitle
\begin{center}
    \vspace{-6mm}
	\centering
	\captionsetup{type=figure}
	\includegraphics[width=.99\textwidth]{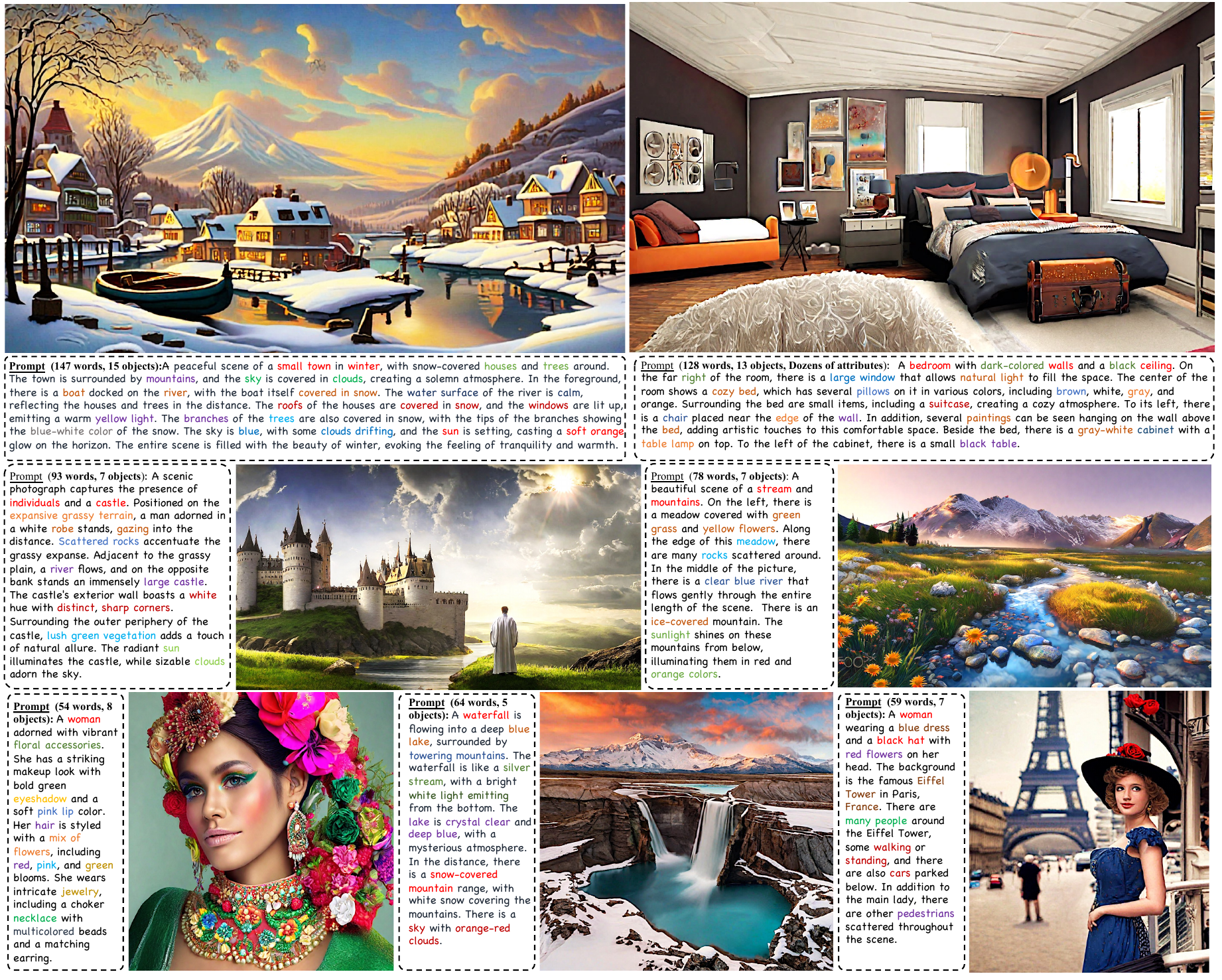}\vspace{-2mm}
    \vspace{-2mm}
    \captionof{figure}{\textbf{
    Examples 
    of 
    Paragraph-Image Alignment from ParaDiffusion.} 
    With the powerful semantic understanding capabilities of the LLM, ParaDiffusion 
    is capable of 
    generating highly \textit{aesthetic} and 
    sophisticated
    images, \textit{aligning well} with 
    long 
    textual content.}
    \label{fig:teaser}
    \vspace{-2mm}
\end{center}
}]

\maketitle
\begin{abstract}
\vspace{-2mm}
Text-to-image models have recently experienced rapid development, achieving 
astonishing
performance in terms of fidelity and textual alignment capabilities.
However, given a long paragraph (up to 512 words), these generation models still struggle to achieve strong alignment and are unable to generate images depicting complex scenes. 
%
In this paper, we introduce an information-enriched diffusion model for paragraph-to-image generation task, 
termed 
ParaDiffusion, 
which 
delves into the transference of the extensive semantic comprehension capabilities of large language models to the task of image generation.
%
At its
core is 
using a large language model (\textit{e.g.,} Llama V2) to encode long-form text, followed by fine-tuning with LoRA to align the text-image feature spaces in the generation task.
%
To facilitate the training of long-text semantic alignment,
we also curated a high-quality paragraph-image pair dataset, namely ParaImage.
This dataset 
contains a 
small amount of 
high-quality, 
meticulously annotated data, and a
large-scale synthetic dataset with long text descriptions being  generated using a vision-language model.
%

%
%

Experiments demonstrate that ParaDiffusion outperforms state-of-the-art models (SD XL, DeepFloyd IF) on ViLG-300 and ParaPrompts, achieving up to $45\%$ human voting rate improvements for text faithfulness. 
Code and data can be found at:
\href{https://github.com/weijiawu/ParaDiffusion}{\color{blue}{\tt ParaDiffusion}}.

\end{abstract}

\section{Introduction}

Recently, text-to-image (T2I) generative models have emerged as focal points of scholarly attention and progress within the computer version field.
Some notable works, such as Stable Diffusion~\cite{rombach2022high} of Stability AI, DALL-E2~\cite{ramesh2022hierarchical} of OpenAI, Imagen~\cite{saharia2022photorealistic} of Google, RAPHAEL~\cite{xue2023raphael} of SenseTime, and Emu~\cite{dai2023emu} of Meta, have achieved substantial progress, accomplishments, and influence for text-based image generation tasks.
%
%
As the name implies, semantic alignment proficiency is paramount for text-guided image generation tasks, wherein the model is required to generate image content corresponding to any provided textual description.
The majority of current T2I models focus on generating high-quality images 
based on processing \textit{relatively short textual inputs and simple descriptions.}
The long-text semantic alignment in the field of text-guided image generation still poses 
a few
challenges.

Given a comprehensive paragraph, as depicted in Figure~\ref{fig:teaser}, extending up to $512$ words, the generation model is required to generate a detailed image that encompasses all the objects, attributes, and spatial positions mentioned within the given paragraph.
For such a task, we define it as \textit{paragraph-to-image generation}.
The majority of current T2I models are unable to handle such intricate long-text semantic alignment, primarily due to 
the two
constraints: 
1) \textbf{Data limitations}. The mainstream public text-image pair dataset, LAION-5B~\cite{schuhmann2022laion}, offers only simple text-image pair information, with an average textual description consisting of approximately $11$ words, which suffer
from a lack of informative content. 
The simplicity of the image-text descriptions limits the model to learn complex and long-text semantic alignment.
%
%
%
2) \textbf{Architecture constraints}. 
Existing T2I models, such as Stable Diffusion and DALL-E2, employ CLIP~\cite{radford2021learning} trained on LAION-5B as 
the 
text encoder, which only supports a maximum of $77$ tokens.
Recently, DeepFloyd \cite{DeepFloyd} and PIXART-$\alpha$ \cite{chen2023pixart} have attempted to leverage the powerful textual understanding capabilities of large language model (\textit{e.g.,} T5-XXL~\cite{raffel2020exploring}) for text-image generation tasks, and demonstrated that T5-XXL outperforms CLIP in image-text alignment.
%
Note that, 
these works merely extended the text embedding of text encoder from $77$ tokens to $128$ tokens without investigating the semantic alignment ability for long-text inputs. 
Furthermore, the T5-XXL model is trained on pure text data without prior knowledge of image-text pairs, and aligning text embeddings of the frozen T5-XXL with visual features in a brute-force manner 
may not be the optimal solution.

In this paper, we explore solutions of long-term text and image alignment 
from the perspectives of 
both
training data and the network structure.

\textbf{For the first challenge,} 
we construct and present a high-quality, textual rich paragraph-to-image pairs dataset, namely ParaImage, where the corresponding textual descriptions can extend up to 400 words.
The long-term text description encompasses the objects, attributes, and spatial locations in the image, along with the corresponding visual style.
ParaImage consists primarily of two types of data: 
1) ParaImage-Big with Generative Captions. We select four million high-quality images from LAION 5B and employ a powerful vision-language model~(\textit{i.e.,} CogVLM~\cite{WeihanWang}) to generate semantically rich textual descriptions. 
This dataset is primarily utilized to achieve alignment between long text and images, enabling the diffusion model to perceive the rich semantic information embedded in lengthy textual descriptions.
2) ParaImage-Small with Manual Captions. 
A few thousand high-quality images are thoughtfully selected from a dataset with $0.6$ million of images with some common principles in photography, then professionally annotated by skilled annotators. 
Considering the inherent limitations in precision associated with synthetic data and the efficiency of quality-tuning~\cite{dai2023emu}, it is necessary to construct a manually annotated, high-quality paragraph-image pair dataset for the final stage of quality tuning.

\textbf{For the second challenge,} 
we explore the transfer of long-text semantic understanding capabilities from the state-of-the-art large language models, \textit{i.e.,} Llama V2~\cite{touvron2023llama}, to the paragraph-to-image generation task.
To harness the robust performance of LLM more effectively, different from the prior methods using frozen weight, we design an efficient training strategy to fine-tune Llama V2 concurrently with the optimization of diffusion models.
This ensures that the extracted text embeddings are more compatible with the text-image pair space.
Besides, the design enables the adaptation of a decoder-only LLM to text-to-image generation tasks.
This, in turn, allows us to leverage the advantages of a decoder-only LLM~(\textit{i.e.,} Llama V2), such as the powerful understanding ability from the larger training text corpus (four times that of T5).
We show 
some generated examples
by
ParaDiffusion in Figure~\ref{fig:teaser}.

Our main contributions are summarized as follows:
\begin{itemize}

    \item For the first time, we explore the long-form text alignment challenge %
    for the
    image generation task, 
    namely, 
    the \textit{paragraph-to-image generation} task.
    A comprehensive solution from both data and algorithmic aspects is provided.
    
    \item We introduce a high-quality, 
    rich-text 
    paragraph-to-image pairs dataset, namely \textit{ParaImage}, where the associated textual descriptions extend up to 400 words, meticulously documenting object identities, properties, spatial relationships, and image style, \textit{etc}.
    
%
    \item We re-evaluate how to better transfer the semantic understanding capabilities of LLM to the text-image generation task, proposing an effective training strategy for fine-tuning LLM, \textit{e.g.,} Llama V2.

    \item Experiments demonstrate that ParaDiffusion outperforms state-of-the-art models (SD XL, DeepFloyd IF) on ViLG-300 and ParaPrompts, achieving up to $45\%$ human voting rate improvements for text faithfulness. 
\end{itemize}

\section{Related Work}
\label{sec:relatedwork}

\textbf{Text-to-Image Models.}
Text-to-image generation   \cite{ho2020denoising} require the model to generate an image corresponding to a given textual description.
Recently, diffusion-based methods \cite{rombach2022high,ramesh2022hierarchical,dai2023emu,yu2022scaling,gu2023mix,he2023ptqd,chang2023muse,kawar2023imagic} have demonstrated remarkable performance and found applications in various downstream tasks, including controllable generation   \cite{zhang2023adding,ruiz2023dreambooth}, controlled editing   \cite{hertz2022prompt}, and perception tasks   \cite{wu2023diffumask,wu2024datasetdm}.
%
Stable Diffusion~\cite{rombach2022high} enhances and accelerates the traditional DDPM   \cite{ho2020denoising} by conducting denoising processes in the latent space. 
Imagen   \cite{saharia2022photorealistic} firstly incorporates large frozen language models~(\textit{i.e.,} T5$-$XXL   \cite{raffel2020exploring}) as text encoders for text-to-image generation tasks, demonstrating their significant performance.
DALL-E2 utilizes CLIP   \cite{radford2021learning} as a text encoder and a diffusion model   \cite{saharia2022image} as a decoder to address text-to-image generation tasks.
Emu   \cite{dai2023emu}, on the other hand, introduces 
new insight that supervised fine-tuning with small but 
high-quality, 
visually appealing images can significantly improve the generation quality.
RAPHAEL   \cite{xue2023raphael}, ERNIE-ViLG   \cite{feng2023ernie}, and Ediffi   \cite{balaji2022ediffi} approach the task of text-to-image generation from the perspective of an ensemble of expert denoisers, exploring potential gains in performance.
Recently, DeepFloyd   \cite{DeepFloyd} and PIXART-$\alpha$   \cite{chen2023pixart} further validate the superior text-image semantic alignment capabilities of large language models over CLIP, as they both employ T5 XXL as a text encoder.
However, these models still solely explore short-text textual alignment tasks, restricting the text encoder to within 128 tokens.
They do not delve deeply into unlocking the full potential of large language models (LLM) and lack exploration of data methodologies. 
In contrast to these approaches, we further investigate the gains in rich paragraph-image alignment capabilities offered by the state-of-the-art language model (\textit{i.e.,}  Llama V2   \cite{touvron2023llama}).
Additionally, we provide an effective solution from a data-centric perspective.

\textbf
{Large Language Models.}
Large language models primarily consist of two mainstream architectures: the Encoder-Decoder architecture   \cite{raffel2020exploring,lewis2019bart} and the decoder-only architecture   \cite{touvron2023llama_v1,touvron2023llama,ouyang2022training}.
Recently, decoder-only models, such as ChatGPT   \cite{ouyang2022training}, GPT-4,  Llama V1   \cite{touvron2023llama_v1}, and  Llama V2   \cite{touvron2023llama}, have exerted significant influence and advancements across various domains.
But for encoder-decoder architecture, as far as we know, the latest one, T5   \cite{raffel2020exploring}, was proposed three years ago.
Decoder-only architectures have achieved significant victories in both training data scale and semantic understanding performance. 
However, existing text-to-image models   \cite{saharia2022photorealistic,DeepFloyd,chen2023pixart}, only conduct exploration of the encoder-decoder architecture, \textit{i.e.,} T5   \cite{raffel2020exploring}, as a text encoder.
This choice is made under the assumption that the decoder-only architecture might not efficiently extract textual features for text-image task.
In this paper, we propose a universal and efficient fine-tuning strategy to adapt any decoder-only architecture~(\textit{e.g.,}  Llama V2   \cite{touvron2023llama}) to a text-guided image generation model.
We present an intriguing insight: directly using frozen LLMs   \cite{saharia2022photorealistic,DeepFloyd} as text encoders is not an elegant solution, while LLMs are trained on pure text data, without considering whether the text embedding is suitable for the text-image feature space.
Inspired by the success of instruction-tuning   \cite{wang2022self,lester2021power} and LoRA   \cite{hu2021lora}, we propose a strategy for paragraph-image alignment learning with language model adaptation. 
This involves freezing the pretrained Large Language Model weights and introducing a certain degree of trainable parameters to learn the textual relationships between paragraphs and images.
%
%
%
%

%

\textbf{LLM-based Layout Generation}
Currently, there is a significant lack of research combining Large Language Models (LLMs) with text-to-image generation models.
Only a few studies~\cite{gani2023llm,lian2023llm,feng2024layoutgpt} have explored the use of LLMs for Layout-to-Image Generation, where these works leverage the LLMs to reason over numerical and spatial concepts in text conditions.
LayoutGPT~\cite{feng2024layoutgpt} proposed to use LLMs as visual planners by generating layouts from text conditions, and thus collaborate with visual generative models.
Similarly, LLM-grounded Diffusion~\cite{lian2023llm} and LLM Blueprint~\cite{gani2023llm} focus on extracting critical components from text prompts using LLMs.
Currently, these approaches suffer from two main drawbacks: 
1) They require multiple models and hand-crafted strategies, making their pipelines complex.
They typically involve using an LLM to generate a layout first, followed by layout-based image generation to produce image. 
2) The layout-based image generation models utilize box-conditional image generation models~\cite{li2023gligen}, which need to be trained on detection datasets.
Therefore, these works can be regarded as extensions of box-guided generation.
In contrast, our model focuses directly on text-guided image generation tasks and only requires training with image-text pairs. Additionally, our pipeline is very simple, as it is an end-to-end trainable pipeline.

\section{Approach}
\label{sec:methodology}

As discussed earlier, we propose a comprehensive solution at both the data and architecture levels for the paragraph-to-image generation task.
Thus, this section is divided into two parts: 1) Algorithm Level~(
\S\ref{Algorithm}). We introduce a paragraph-to-image generation architecture with diffusion model.
%
%
2) Dataset Level~(
\S\ref{Dataset}). We present a high-quality, textual rich paragraph-to-image pairs dataset, where the corresponding textual descriptions can extend up to $400$ words.

\subsection{Algorithm: ParaDiffusion}
\label{Algorithm}

Following LDM \cite{rombach2022high}, our architecture consists of three components: a text encoder to encode textual descriptions, an autoencoder (AE) to encode an image into latent embeddings, and a U-Net~\cite{ronneberger2015u} to learn the denoising process.

\begin{figure*}[t]
    \begin{center}
        \includegraphics[width=.98\linewidth]{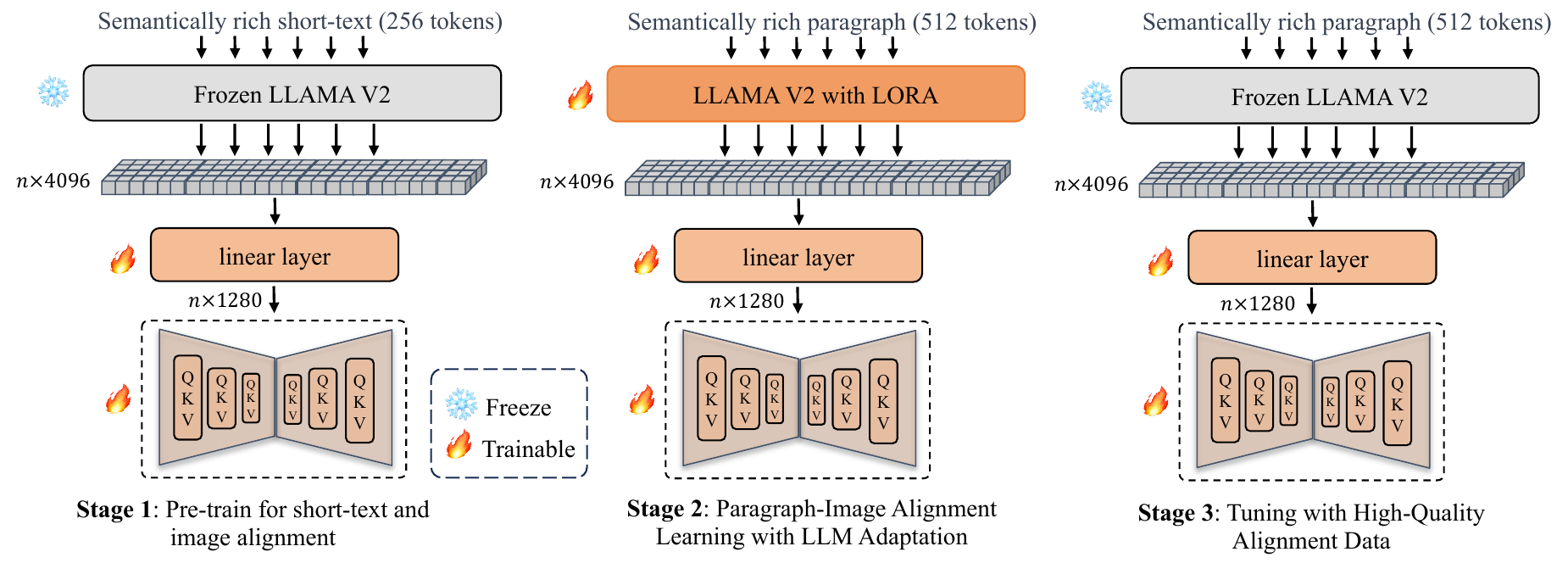}
    \end{center}
    \vspace{-0.6cm}
    \caption{\textbf{Pipeline of Methodology}. 
    The training pipeline of ParaDiffusion mainly includes three stages: 1) Stage-1 for pretraining is based on 0.3 billion samples to acquire general text-image knowledge. 
    2) Stage-2 employ millions of data to simultaneously fine-tune LLM and the diffusion model for Paragraph-Image Alignment. 
    3) Quality tuning with curated high-quality annotated data~(\textit{i.e.,} ParaImage-Small). }
    \label{fig:pipeline}
    \vspace{-0.2cm}
\end{figure*}

\subsubsection{Text Paragraph Encoder.}
Prior works   \cite{DeepFloyd,chen2023pixart} directly utilize \textit{frozen} encoder-decoder language model~(\textit{i.e.,} T5   \cite{raffel2020exploring}) for short-text to image generation tasks, restricting the text encoder to within 128 tokens.
We present an insight that efficiently fine-tuning a more powerful decoder-only language model can yield stronger performance in long-text alignment (up to 512 tokens).
%
%
Compared to the encoder-decoder architecture of large language models, recent decoder-only models, such as ChatGPT   \cite{ouyang2022training}, and Llama  V2   \cite{touvron2023llama}, have garnered more attention and success.
There are some evident advantages to using the decoder-only language models as text encoders:
1) Models like Llama  and the GPT series, based on the decoder-only architecture, exhibit stronger semantic understanding and generalization capabilities.
2) The training corpora are also nearly four times larger than that of the encoder-decoder language model, \textit{i.e.,} T5   \cite{raffel2020exploring}.
However, typically, decoder-only architectures are not adept at feature extraction and mapping tasks. 
Therefore, we propose Paragraph-Image Alignment Learning with Language Model Adaptation to address this issue, 
adapting the decoder-only model, \textit{i.e.,} Llama  V2   \cite{touvron2023llama} for text-to-image generation tasks, as shown in Figure~\ref{fig:pipeline}.
%


\subsubsection{Paragraph-Image Alignment Learning with Language Model Adaptation.}
\label{alignmentlearning}

Given a text description $y$ and image $x$, the standard Text to Image Diffusion Models is probabilistic model designed to model conditional distributions of the form $p(z \vert y)$,
where a conditional denoising autoencoder \\$\model(\zt{t},t,y);\, t\in{\{1, \dots, T\}}$ is used to learn
the reverse process of a fixed Markov Chain of length $T$.
The corresponding objective can be simplified to:
\begin{equation}
\expec_{\encoder(x), y, \epsilon \sim \mathcal{N}(0, 1), t }\Big[ \Vert \epsilon - \model(z_{t},t, \conditioner(y)) \Vert_{2}^{2}\Big] \, ,
\label{eq:cond_loss}
\end{equation}
where $\conditioner$ denotes the text encoder and its output is a text embedding ($n \times 1280$), as shown in Figure~\ref{fig:pipeline}. $\encoder$ refers to a AE   \cite{rombach2022high} for mapping image to latent feature.
Typically, during training, we optimize only the Unet of the diffusion model $\model$, while freezing the $\conditioner$. 
This is because we consider the text embedding from $\conditioner$, \textit{i.e.,} CLIP, to be well-suited for text-image alignment tasks.
However, when using an LLM as the text encoder, it is essential to consider whether a frozen LLM is appropriate for this context.

Inspired by the success of instruction-tuning   \cite{wang2022self,lester2021power} and LoRA   \cite{hu2021lora}, we propose a strategy for paragraph-image alignment learning with language model adaptation. 
This involves freezing the pretrained Large Language Model weights $\tau_{\theta_0}$ and introducing a certain degree of trainable parameters $\Delta\theta(\Theta)$ to learn the textual relationships between paragraphs and images.
The objective is revised as follows:
\begin{equation}
\expec_{\encoder(x), y, \epsilon \sim \mathcal{N}(0, 1), t }\Big[ \Vert \epsilon - \model(z_{t},t, \tau_{\theta_0+\Delta\theta(\Theta)}(y)) \Vert_{2}^{2}\Big] \, ,
\label{eq:cond_loss_1}
\end{equation}
where 
both $\model$ and $\tau_{\Delta\theta(\Theta)}$ can be jointly optimized for learning better text representation.
Compared to direct fine-tuning, this strategy offers two advantages: 
1) During the training of paragraph-image alignment, it preserves the powerful semantic understanding capabilities of the LLM, preventing knowledge overfitting to simple text-image semantics~\cite{hu2021lora}.
2) Storage and compute efficient.
Requires only limited computational resources and incurs no additional inference costs.
%

%

\begin{figure*}[t]
    \begin{center}
        \includegraphics[width=.98\linewidth]{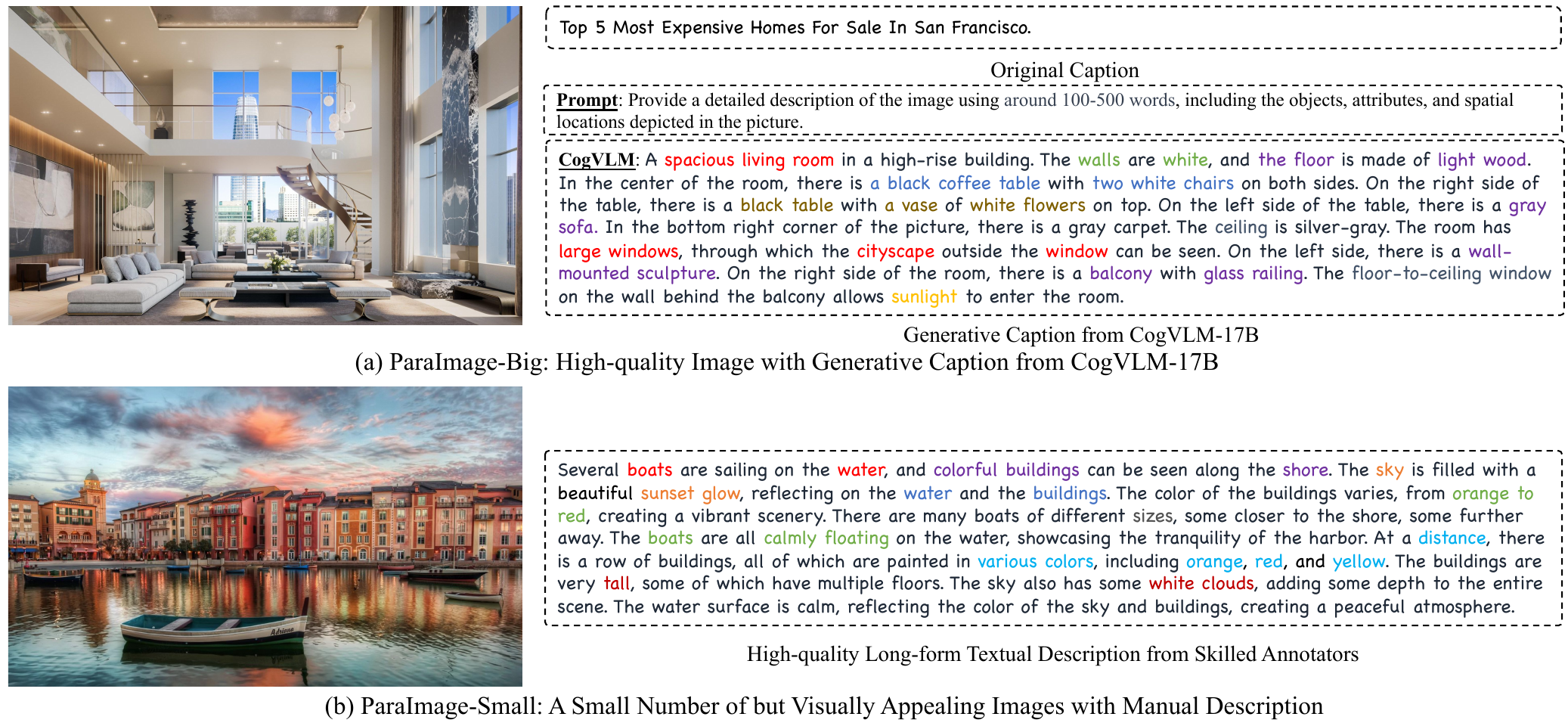}
    \end{center}
    \vspace{-0.6cm}
    \caption{\textbf{Examples of the proposed ParaImage dataset}. (a) High-quality images with generative captions~(ParaImage-Big) are primarily employed for the paragraph-image alignment learning in Stage 2. (b) Aesthetic images with manual long-term description~(ParaImage-Small) are primarily used for quality-tuning in Stage 3.}
    \label{fig:examples}
    \vspace{-0.4cm}
\end{figure*}

\subsubsection{Training Strategy.}
As shown in Figure~\ref{fig:pipeline}, ParaDiffusion adopts a three-stage training approach.
%
%
Similar to the prior works   \cite{dai2023emu,saharia2022photorealistic}, stage 1 is employed to acquire general text-image semantic alignment knowledge.
Stage 2 is introduced to simultaneously fine-tune the LLM and the diffusion model for paragraph-image alignment.
%
%
%
In Stage 3, a high-quality small dataset, consisting of $3k$ carefully selected images, is used to further enhance the model performance.
%

\textbf{Stage 1: Pre-train for Short-text and Image Alignment.}
In this stage, we gathered a dataset of 300 million examples to train the Unet of diffusion model with 1.3 billion parameters.
Among these, $100$ million examples were selected from the 5 billion images of the LAION dataset   \cite{schuhmann2022laion}.
Additionally, we constructed an internal dataset containing over $200$ million examples.
Similar to SDXL~\cite{podell2023sdxl}, the model was trained with progressively increasing resolutions. 
It started with training at a resolution of 256 and was later adjusted to a resolution of 512.
The entire pre-training process took 5 days on 56 A100 GPUs.

\textbf{Stage 2: Paragraph-Image Alignment Learning with LLM Adaptation.}
To perform paragraph-image alignment learning~(\S~\ref{alignmentlearning}), we construct a large scale dataset with several million paragraph-image pairs, namely ParaImage-Big, where the long-form text is generated by CogVLM   \cite{WeihanWang}, as illustrated in Figure~\ref{fig:examples}(a).
The entire paragraph-image alignment learning process took 2 days on 56 A100 GPUs.


\textbf{Stage 3: High-Quality Alignment Data.}
%
%
Finally, we created an extremely small but high-quality dataset, namely ParaImage-Small, to further enhance the performance of model with a small learning rate.
This dataset consists of 3k manually selected images, each accompanied by human-annotated long-text descriptions. 
These descriptions provide detailed information about the objects in the images, including their attributes and spatial relationships.
%


%

\subsection{Dataset: ParaImage}
\label{Dataset}


\subsubsection{Image with Generative Captions.}
\label{generativecaption}
\begin{figure}[tbp]
    \begin{center}
    \includegraphics[width=.98\linewidth]{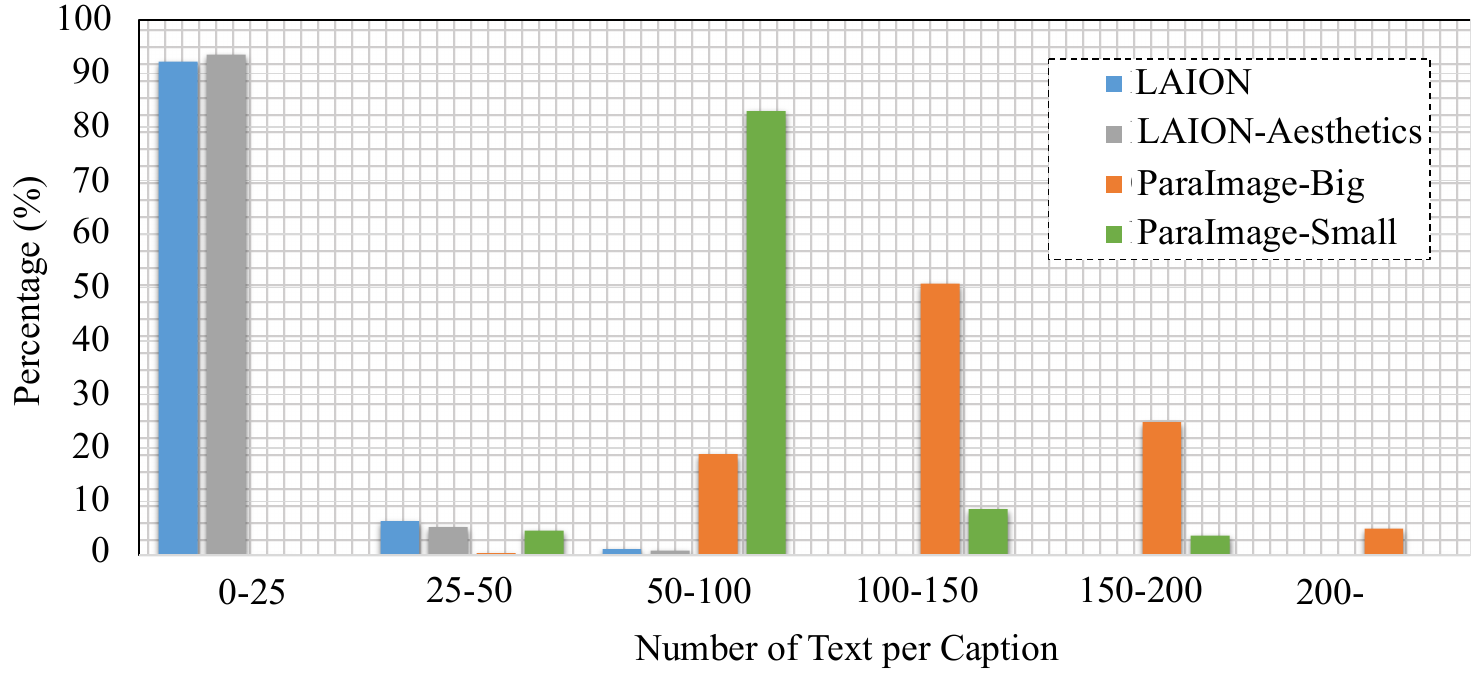}
    \end{center}
    \caption{\textbf{Distribution of Caption Length.} The textual descriptions of the proposed dataset~(ParaImage) far exceed those of currently available public datasets.}
    \label{fig:Caption}
\end{figure}

Inspired by the success of current large vision-language models   \cite{WeihanWang,zhu2023minigpt,liu2023visual}, we propose to leverage the SOTA model, \textit{i.e.,} CogVLM for automatically generating extensive long-form textual annotations for images. 
Firstly, we gathered approximately 3.3 million high-quality images from the LAION-Aesthetics   \cite{laion_aesthetics} and SAM datasets~\cite{kirillov2023segment}.
For the LAION-Aesthetics, we downloaded around $8$ million images with aesthetics scores above $6$ and then filtered them to a set of $3.2$ million images with a minimum short edge resolution of 512 pixels.
For SAM dataset, we obtained 2 million images and further filtered out those with mosaic elements, resulting in a final dataset of around 100k images.
Then, we prompted CogVLM to generate corresponding long-form textual descriptions for the images, as illustrated in Figure~\ref{fig:examples}(a).
Figure~\ref{fig:Caption} and Table~\ref{table_dataset} present a detailed statistical comparison, and it is evident that in terms of textual descriptions, the captions of the proposed dataset are richer and longer in semantic content. 
$92\%$ of captions from LAION have a text length of fewer than 25 words, whereas over $70\%$ of captions from ParaImage-Big exceed $100$ words, with a few extending to over 200 words.

\subsubsection{ParaImage-Small: Image with Manual Captions.}
\label{manualcaptions}
The generated captions from CogVLM cannot be guaranteed to be $100\%$ accurate; therefore, it is essential to create high-quality dataset with manual annotations~(\textit{i.e.,} ParaImage-Small).
This dataset consists of $3$k, manually selected by annotators from $650k$ high-quality images of LAION-Aesthetics, adhering to common principles in photography.
Then, the $3k$ images were annotated with long-text descriptions, detailing the objects in the images, the attributes of these objects, and spatial relationships.
%
%
The details for \textit{Description Annotation Rule} are as follows:

\begin{itemize}
    \item The description should be \textbf{no less than $50$ words and no more than $500$ words}.
    \item The first sentence of the description should commence with a portrayal of the main subject of the image, \textit{e.g.,} A close-up photo of a beautiful woman.
    \item The description should be objective and avoid subjective emotions and speculations, such as (`On a cloudy day, it might rain in the future').
    \item The description should cover details about the objects' number, color, appearance, and location.
    \item The description should follow a logical sequence, such as from left to right or from the center outward.
    \item Avoid using prefixes like `In the picture is,' `This image depicts', or `Here is.'
\end{itemize}

\begin{table*}[t]
    \centering
    \small 
    \setlength{\tabcolsep}{1mm}
    \caption{\textbf{Comparison of Data Statistics.} `short side' denotes average length of short side for input image.
    }
    \scriptsize
\begin{tabular}{l|cc|cc}
    \multirow{2}{*}{Dataset}  &  \multicolumn{2}{c|}{Image}&  \multicolumn{2}{c}{Caption}\\
    & Number  & Short Side & Average Words & Average Nouns  \\
    \shline
    \hline
    LAION~\cite{schuhmann2022laion}& 2.3b & 537.2 & 11.8 &6.4 \\
    LAION-Aesthetics~\cite{laion_aesthetics}  & 625k & 493.6 & 11.3 &6.8 \\
    \hline
    ParaImage-Big&  3.3m & 771.3 & 132.9 & 46.8\\
    ParaImage-Small& 3.1k & 1326.2 & 70.6  & 34.2  \\
    \end{tabular}
    
    \label{table_dataset}
\end{table*}

Details regarding the selection principles of images are provided in the supplementary materials.
Based on the images selection annotation rules, we invited and trained $10$ annotators to score the images from $1$ to $5$ based on aesthetics and content, with $5$ being the highest score. 
%
%
There 
are 
two rounds of quality inspection, each round involving $3$ staff members for review.
Finally, we selected images with the highest score, resulting in around $3000$ images.
With these images, we 
hired 
$20$ annotators to provide a detailed description for each image following the annotation guidelines above.
%
%
Then two rounds of quality audits were conducted afterwards, with two people in each round. 
For the images that does not meet the evaluation criteria, we would conduct revisions. 
The entire labeling process took two weeks, with one week for selecting pictures and the other week to provide detailed text descriptions.

\section{Experiments}
\label{sec:experiments}

\begin{figure}[tbp]
    \begin{center}
        \includegraphics[width=.98\linewidth]{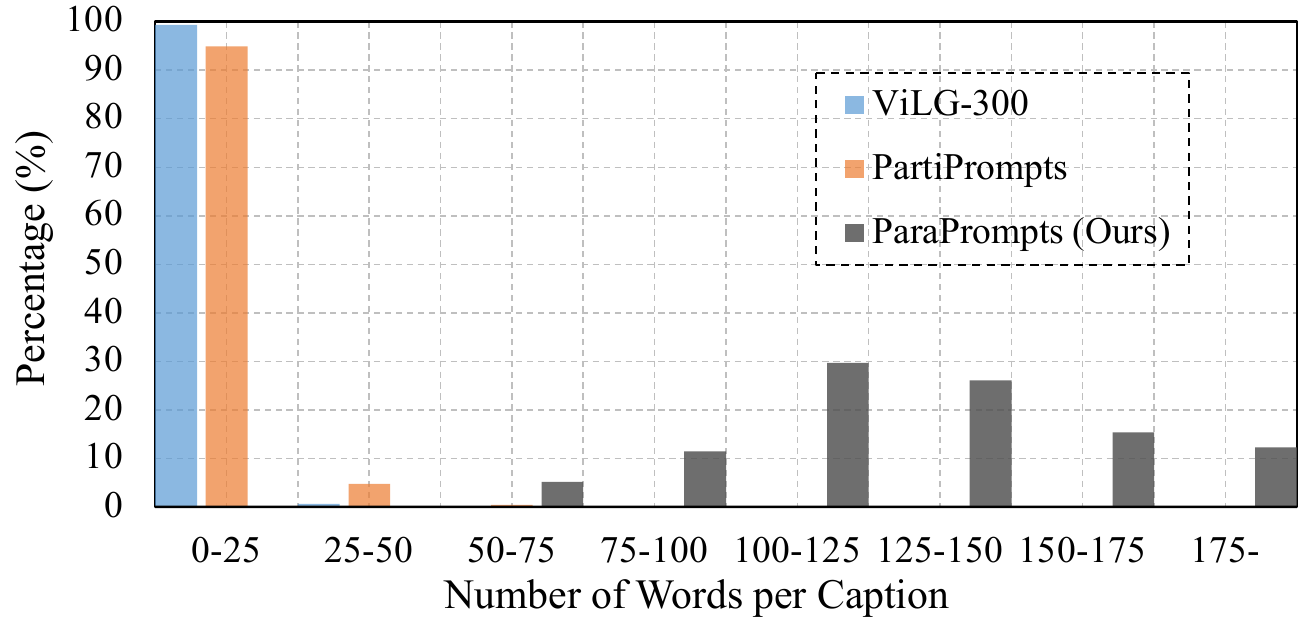}
    \end{center}
    \caption{\textbf{Distribution of Caption Length for Different Evaluation Dataset.} Our ParaPrompts dataset offers a high proportion of long-text descriptions. }
    \label{fig:eval}
\end{figure}

\begin{figure}[tbp]
    \begin{center}
        \includegraphics[width=.78\linewidth]{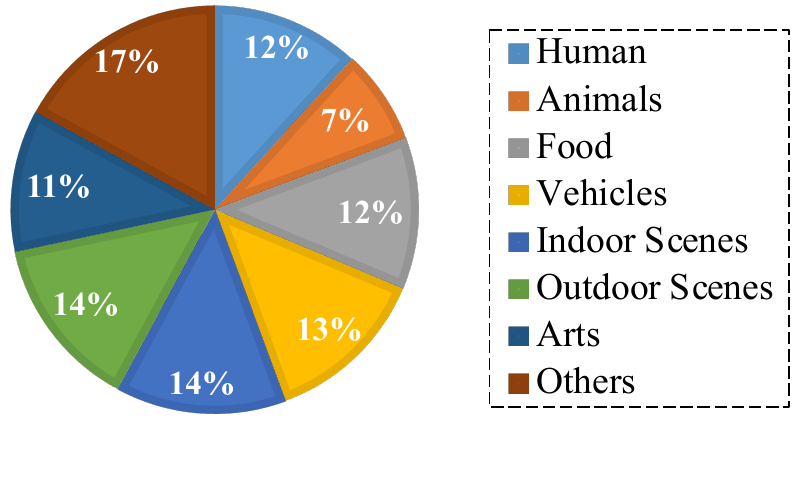}
    \end{center}
    \caption{\textbf{Distribution of Concepts for ParaPrompts.} Our prompts covers the common concepts in human-related activities.}
    \label{fig:eval_pie}
\end{figure}



\subsection{Implementation Details}
\textbf{Algorithm.} Different from the prior works~\cite{rombach2022high,ramesh2022hierarchical,dai2023emu}, for the first time, we used a powerful decoder-only large language model~(\textit{i.e.,} 7B Llama V2~\cite{touvron2023llama}) as the text encoder. 
To learn the long-term paragraph-image alignment, we adjusted the length of extracted text tokens to $512$ tokens, enabling precise alignment for very complex and semantically rich prompts.
For the diffusion model, we follow SDXL~\cite{podell2023sdxl} and use  a Unet 
of 
$1.3$B parameters  to learn the latent Gaussian noise.
Compared to previous work, our training cost is also 
significantly lower, 
with the entire model training process only requiring \textit{$56$ A100 GPUs for $8$ days}.
Following LoRA ~\cite{hu2021lora}, we apply LoRA  to each layer of Llama V2~\cite{touvron2023llama}.
At different stages, we employ varying learning rate strategies.
For Stage 1, the learning rate is set to $8e\mbox{--}5$ with cosine annealing decay for a single cycle. For Stage 2 and Stage 3, the learning rate is $5e\mbox{--}5$ and $1e\mbox{--}5$, respectively.

\textbf{Datasets and Metrics.}
\label{metric}
We 
use
Zero-shot FID-30K on MS-COCO $256\times 256$ \cite{lin2014microsoft} and human evaluation on ViLG-300\cite{feng2023ernie} to assess our algorithm. 
\textit{ParaPrompts.}
Additionally, considering that the current test prompts focus on short text-to-image generation, ignoring the evaluation for paragraph-to-image generation, we introduced a new evaluation set of prompts called ParaPrompts, including $400$ long-text descriptions.
Figure~\ref{fig:eval} illustrates the distribution comparison of prompt lengths between ParaPrompts and the previous test set. 
It is evident that previous prompts testing was mostly concentrated on text alignments within the range of 0-25 words, while our prompts extend to long-text alignments of $100$ words or more. 
Additionally, we present a new insight that longer text descriptions are more challenging, both in terms of visual appeal and text faithfulness.
Relevant discussions and comparisons are provided in the supplementary materials.
Figure~\ref{fig:eval_pie} presents the distribution of prompt content, where we categorized prompts into eight scene categories to achieve a more comprehensive evaluation.
\textit{Text Faithfulness of Sentence Level.} To further evaluate the text faithfulness performance, we propose a more fine-grained metric, where caption of ParaPrompts-400 are divided into $3,945$ sentence fragments. 
Then, an open-domain grounding model ~\cite{lai2023lisa} is employed to segment the related content for each sentence.
Finally, two evaluators manually determine the precision (whether the relevant content is present in the synthesized image).

\subsection{Performance Comparisons and Analysis}

\subsubsection{Fidelity Assessment on COCO Dataset.}
Zero-shot FID-30K on MS-COCO $256\times 256$~\cite{lin2014microsoft} is a universal evaluation method for text-image generation tasks.
Table~\ref{coco} presents relevant comparisons between ParaDiffusion and other existing works.
Our ParaDiffusion achieved an FID score of $9.64$, demonstrating similar performance to PIXART-$\alpha$~\cite{chen2023pixart}.
In comparison, RAPHAEL~\cite{xue2023raphael} and DeepFloyd-IF~\cite{DeepFloyd} achieved better scores, while utilizing larger models with more parameters. 
Furthermore, we would like to argue that FID may not be an appropriate metric for image quality evaluation, where a higher score does not necessarily indicate better-generated images.
Many studies~\cite{chen2023pixart,DeepFloyd,podell2023sdxl}, have demonstrated that, instead, the evaluation by human users is a more authoritative measure.

\begin{table*}[t]
    \centering
    \small 
    \setlength{\tabcolsep}{1mm}
    \caption{\textbf{Performance Comparison on MS-COCO $256\times 256$~\cite{lin2014microsoft} using zero-shot FID-30K.} `\#Params' refers to the parameters of Unet.
    }
    \vspace{-1mm}
    \scriptsize
\begin{tabular}{l| l ccc}
    Method  &  Text Encoder & \#Params &  
     FID-30K$\downarrow$  &  
     Venue/Date
     \\
    \shline
    \hline
    DALL-E & - & 12.0B & 27.50 & Blog, Jan. 2021 \\
    GLIDE~\cite{nichol2021glide} & - & 5.0B &  12.24 & ICML’22 \\
    
    DALL-E2~\cite{ramesh2022hierarchical} &  CLIP~\cite{radford2021learning} & 6.5B &  10.39 & 
     arXiv, April 2022\\
    PIXART-$\alpha$~\cite{chen2023pixart} &  T5$\mbox{-}$XXL~\cite{raffel2020exploring} & 0.6B &   10.65 & arXiv, Oct. 2023\\
    
    SD XL~\cite{podell2023sdxl} &  CLIP~\cite{radford2021learning} &  2.6B &  - & arXiv, Jul. 2023 \\
    GigaGAN~\cite{kang2023scaling} &  CLIP~\cite{radford2021learning} &  0.9B &  9.09 &  CVPR’23\\

    SD~\cite{rombach2022high} & CLIP~\cite{radford2021learning} & 0.9B &  8.32 & CVPR’22 \\
    Imagen~\cite{saharia2022photorealistic} &  T5$\mbox{-}$XXL~\cite{raffel2020exploring} &   3.0B &   7.27 &  NeurIPS’22\\
    
    ERNIE-ViLG 2.0 & CLIP~\cite{radford2021learning} & 22B & 6.75 & CVPR’23\\
    
    DeepFloyd-IF~\cite{DeepFloyd} &  T5$\mbox{-}$XL~\cite{raffel2020exploring} &   4.3B &    6.66 &  Product, May 2023 \\
    RAPHAEL~\cite{xue2023raphael} &  CLIP~\cite{radford2021learning} &   3.0B &    6.61 &  arXiv, May 2023\\
    \hline
    ParaDiffusion &  Llama  V2~\cite{touvron2023llama} &   1.3B &    9.64 &  -\\
    
    \end{tabular}
    
    \label{coco}
\end{table*}

\begin{figure}[tbp]
    \begin{center}
        \includegraphics[width=.98\linewidth]{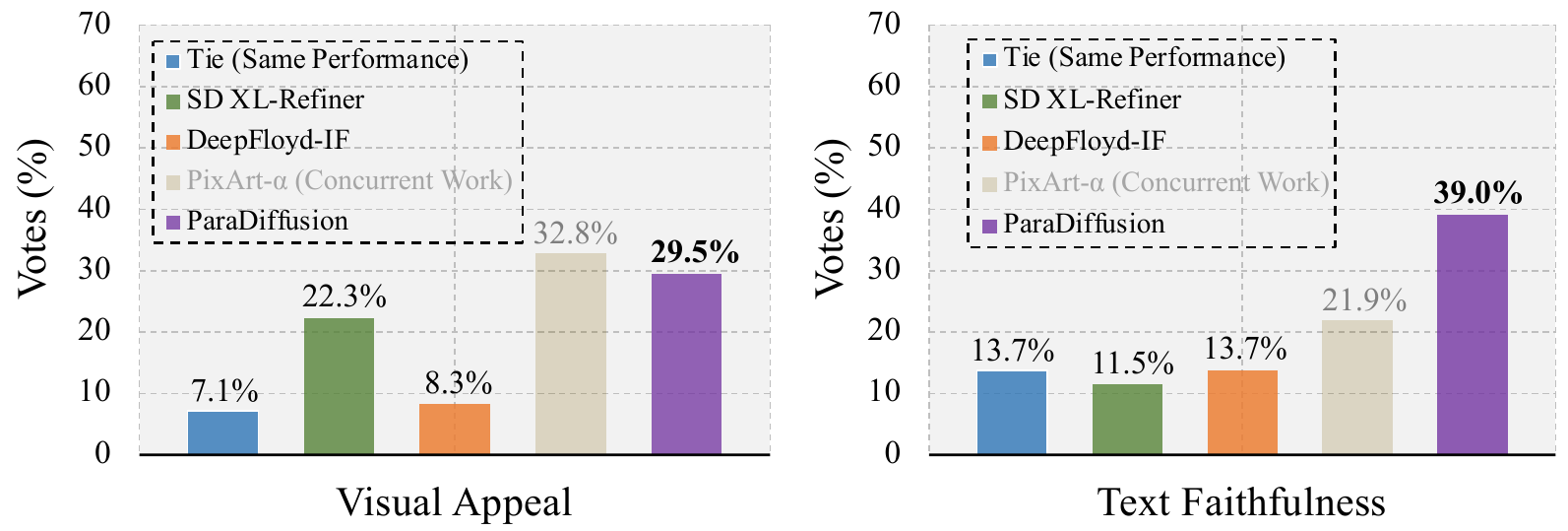}
    \end{center}
    \caption{\textbf{User study on 300 prompts of ViLG-300~\cite{feng2023ernie}.} `Tie' indicates that the image quality of the four models appears similar. }
    \label{fig:ViLG}
\end{figure}

\begin{figure}[tbp]
    \begin{center}
        \includegraphics[width=.98\linewidth]{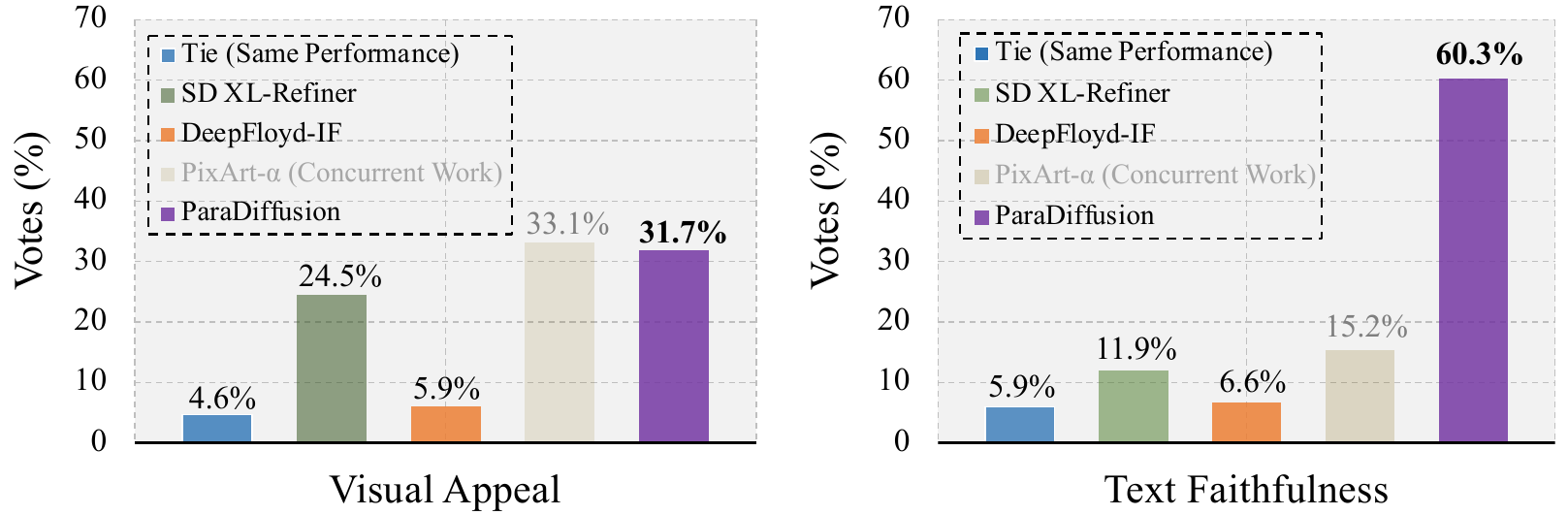}
    \end{center}
    \caption{\textbf{User study on 400 prompts of ParaPrompts.} We only selected open-source models for comparison, when the results of closed-source models~\cite{xue2023raphael,radford2021learning} were unavailable or API calls failed.}
    \label{fig:ParaPrompts}
\end{figure}

\begin{figure}[tbp]
    \begin{center}
        \includegraphics[width=.78\linewidth]{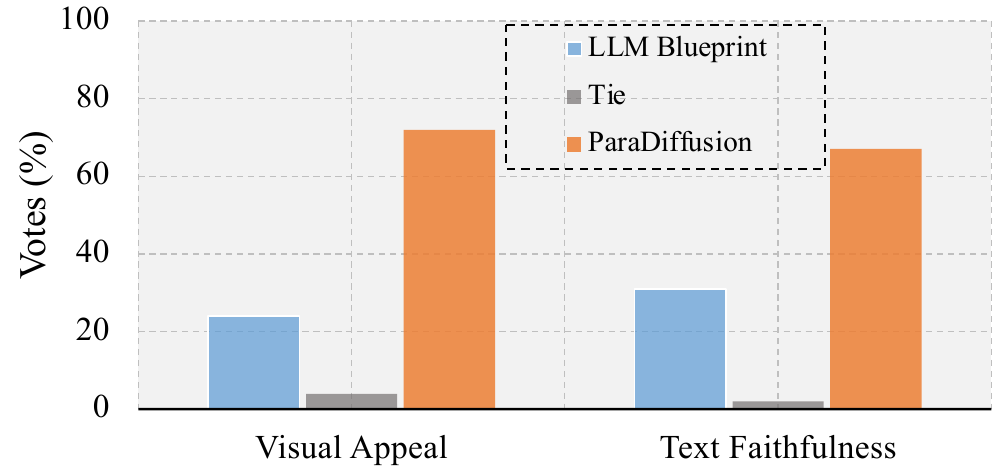}
    \end{center}
    \caption{\textbf{User study for LLM Blueprint~\cite{gani2023llm} and ParaDiffusion on 400 prompts of ParaPrompts.} `Tie' indicates that the image quality of the four models appears similar. }
    \label{fig:layout}
\end{figure}

\begin{figure*}[tbp]
    \begin{center}
        \includegraphics[width=.98\linewidth]{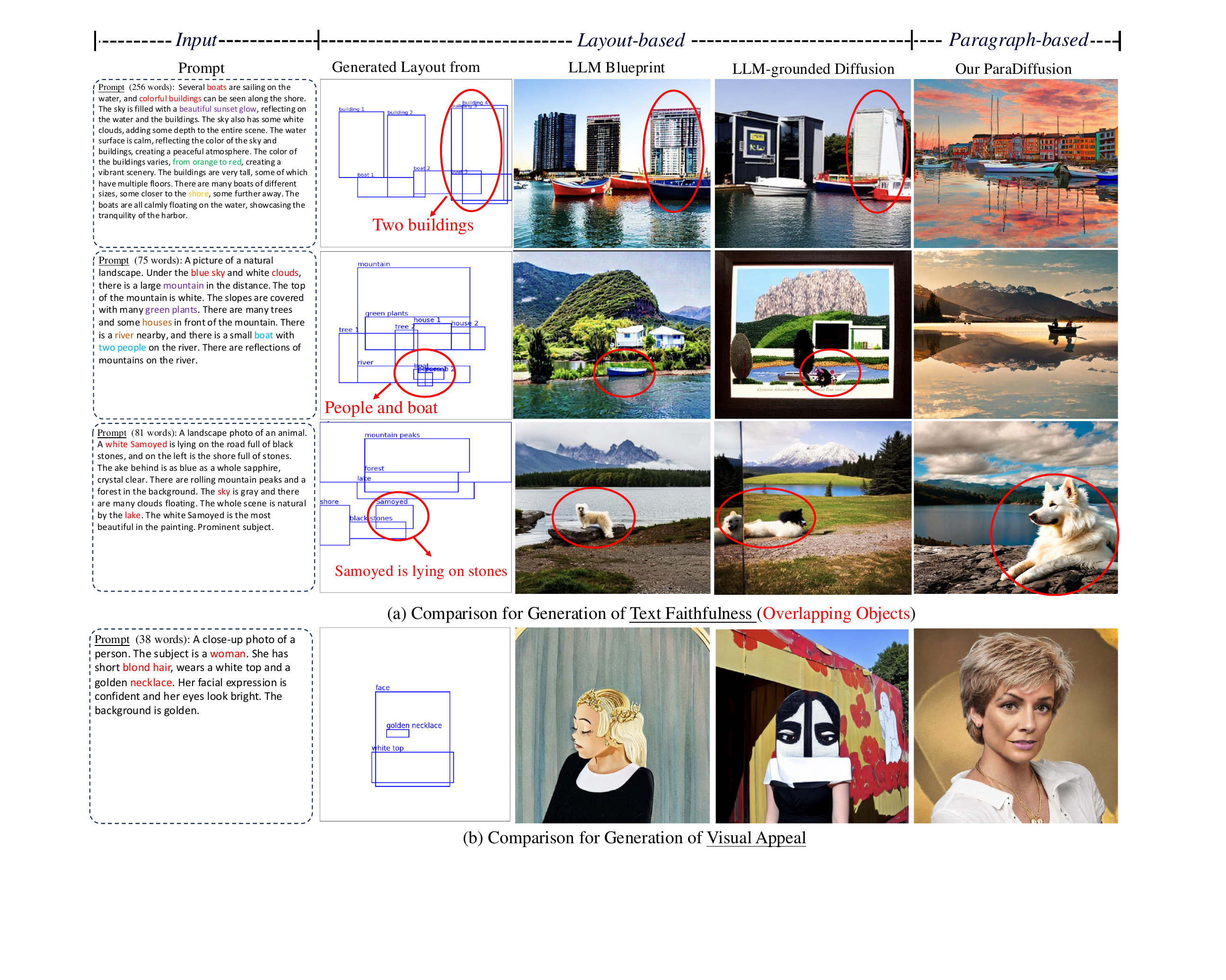}
    \end{center}
    \caption{\textbf{More Visualization Comparison for Layout-based models~(LLM Blueprint~\cite{gani2023llm} and LLM-grounded Diffusion~\cite{lian2023llm}) and ParaDiffusion.} Compared to layout-based methods, our ParaDiffusion achieves better text faithfulness, especially in generating overlapping objects.}
    \label{fig:comparison_layout}
\end{figure*}

\begin{figure}[tbp]
    \begin{center}
        \includegraphics[width=.98\linewidth]{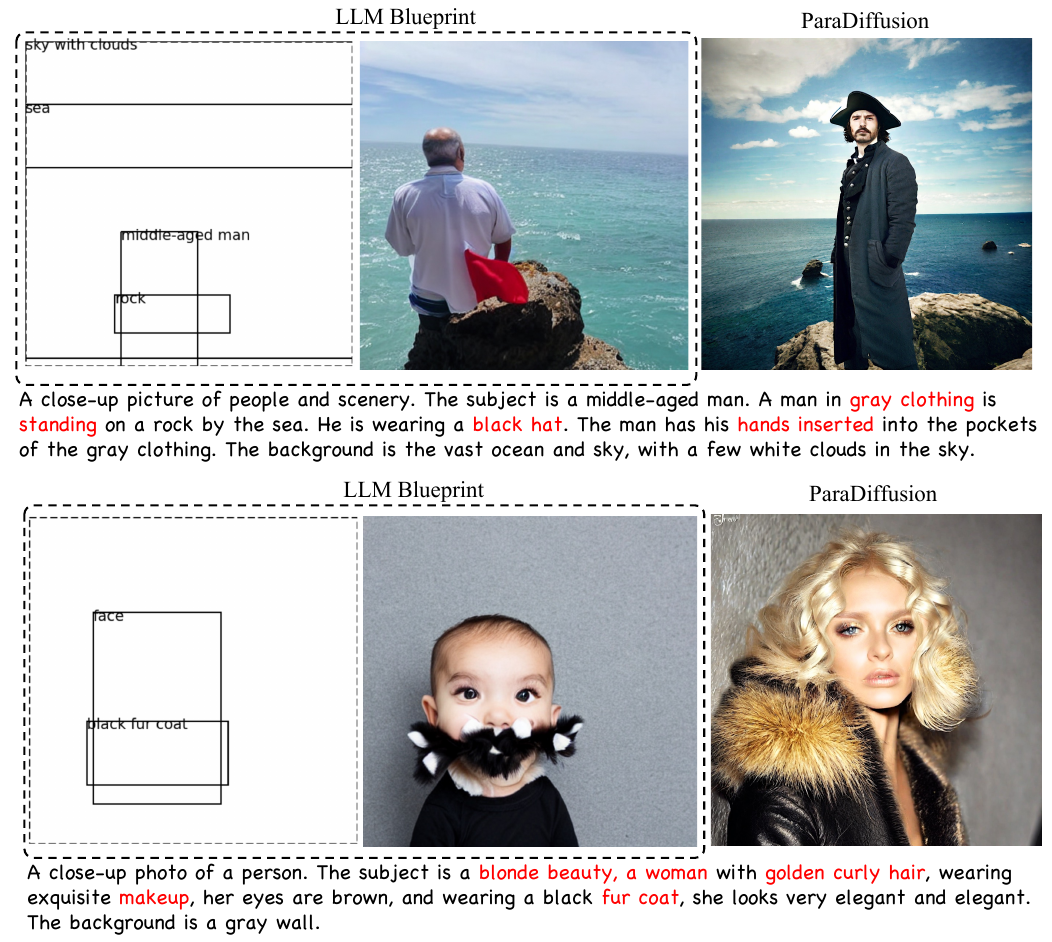}
    \end{center}
    \caption{\textbf{Visualization Comparison for LLM Blueprint~\cite{gani2023llm} and ParaDiffusion.} The effect of ParaDiffusion is better than LLM Blueprint in terms of both Visual Appeal and Text Faithfulness.}
    \label{fig:vis_layout}
\end{figure}

\subsubsection{ViLG-300~\cite{feng2023ernie}.}

Following the prior works~\cite{chen2023pixart,feng2023ernie}, we also conducted a User Study evaluation from the perspectives of visual appeal and text faithfulness on 300 prompts from ViLG-300~\cite{feng2023ernie}.
We only chose the recent models with sota performance for comparative evaluation, as involving human evaluators can be time-consuming.
Figure~\ref{fig:ViLG} presents the related rating proportions from two expert evaluators.
%
%
In cases where there was ambiguity in the voting, the two evaluators discussed and reached a consensus.
\textbf{Visual Appeal.} Our model significantly outperforms DeepFloyd-IF and SD XL-Refiner in terms of visual appeal, while both of these are considered among the best models of the past year.
In comparison to PIXART-$\alpha$\cite{chen2023pixart}, our model has achieved competitive performance.
We want to emphasize that our model did not specifically focus on Visual Appeal. 
Additionally, PIXART-$\alpha$\cite{chen2023pixart} and our work are concurrent efforts. Therefore, we believe the performance is acceptable.
\textbf{Text Faithfulness.}
Figure~\ref{fig:ViLG} (right) shows that our model achieved outstanding performance in Text Faithfulness among the four models, with a voting percentage of $13.7\%$.
`Tie (Same Performance)' received a very high voting percentage.
This is because ViLG-300 includes many simple prompts, leading to consistently good results and making it challenging to differentiate among them.
We provide additional cases in the supplementary material to further analyze this situation.

\begin{table}[t]
    \centering
    \small 
    \setlength{\tabcolsep}{1mm}
    \caption{A fine-grained metric for the precision of sentence fragments (Text Faithfulness) using LISA~\cite{lai2023lisa} on ParaPrompts-400.} 
    \scriptsize
\begin{tabular}{l|cccc}

    Methods  & SD XL  & DeepFloyd-IF  & 
    PixArt-$\alpha$ & ParaDiffusion
    \\
    \shline
    \hline
    Precision & $25.8\%$ & $27.9\%$ & $29.4\%$ & 
    $51.3\%$  

    \end{tabular}
    \label{ablation_33}
\end{table}

\begin{table}[t]
    \centering
    \small 
    \setlength{\tabcolsep}{1mm}
    \caption{\textbf{ Evaluation for Objects, Attributes, and Locations on ParaPrompts 400.} We randomly sample 50 samples from ParaPrompts 400 and conduct a 1-5 scoring assessment, with 5 being the highest score.}
    \scriptsize
\begin{tabular}{l|ccc}

    Methods  & Objects  & Attributes  & 
    Locations 
    \\
    \shline
    \hline
    LLM Blueprint~\cite{gani2023llm} & 3.4 & 3.1 & \textbf{4.2}  \\
    ParaDiffusion & \textbf{3.9} & \textbf{4.1} & 4.0
    \end{tabular}
    \label{parts}
\end{table}


\begin{figure*}[t]
    \begin{center}
        \includegraphics[width=.98\linewidth]{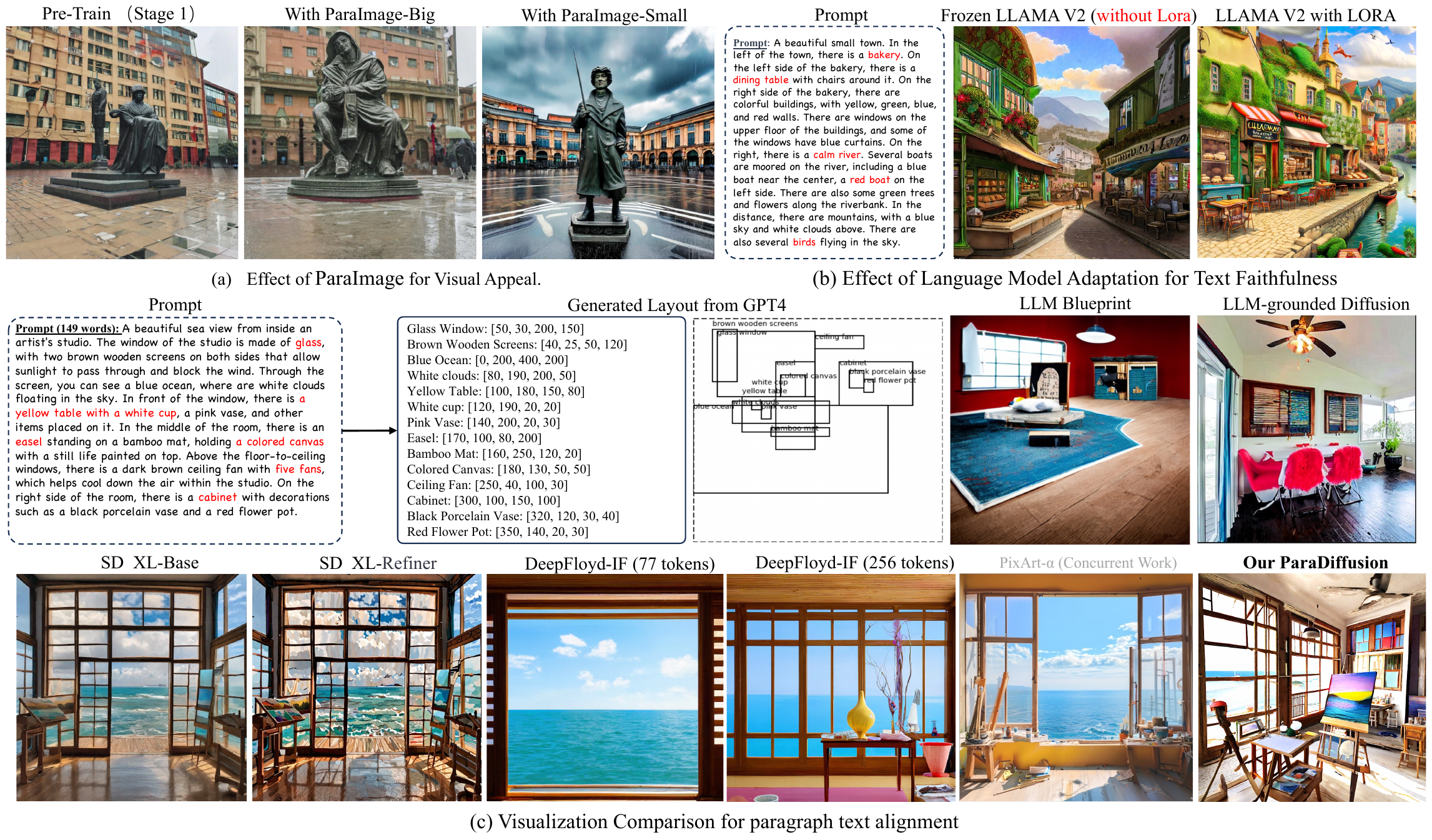}
    \end{center}
    \caption{\textbf{Visualization Comparison.} ParaDiffusion demonstrates exceptional superiority in long-term text alignment.}
    \label{fig:ablation}
\end{figure*}

\begin{figure}[tbp]
    \begin{center}
        \includegraphics[width=.92\linewidth]{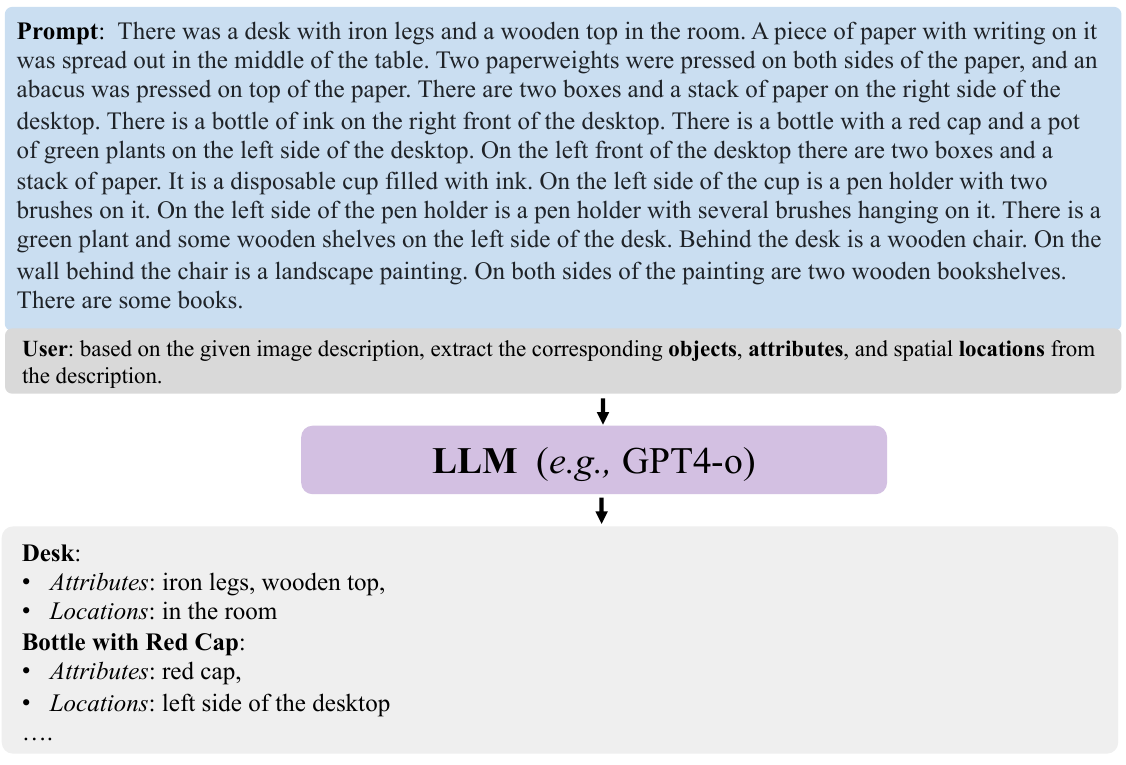}
    \end{center}
    \caption{\textbf{Prompt workflow for detailed concepts in image descriptions.}}
    \label{pipeline-shot}
\end{figure}



\subsubsection{ParaPrompts-400.}
Figure~\ref{fig:ParaPrompts} presents the results on ParaPrompts-400.
\textbf{Visual Appeal.}
Similar to ViLG-300~\cite{feng2023ernie}, our ParaDiffusion achieved outstanding results in terms of visual appeal on the ParaPrompts dataset, surpassing SD XL and DeepFloyd-IF and approaching PIXART-$\alpha$~\cite{chen2023pixart}. 
Compared to the performance on the ViLG-300~\cite{feng2023ernie} dataset, there is a decrease in the voting percentage for the `Tie' category. 
This indicates that in more challenging long-text scenarios, the performance differences among different models become more pronounced.
\textbf{Text Faithfulness.} 
As for paragraph-guided image generation setting, ParaDiffusion demonstrates a more pronounced advantage in text faithfulness, reaching $60.3\%$, while other models achieve only around $10\%$.
This indicates that ParaDiffusion significantly outperforms other models in aligning long-text aspects such as object, object attributes, and object positional relationships.
In the supplementary material, we provide additional visualizations and analyses to support this claim.
%
%
Additionally, we also provide a more fine-grained metric (refer to Section~\ref{metric}), as illustrated in Table~\ref{ablation_33}.
It can be observed that even at the sentence level, our model achieves an absolute lead in text faithfulness alignment, reaching $51.3\%$, surpassing the other models by nearly $20\%$.

\begin{figure*}[t]
    \begin{center}
        \includegraphics[width=.98\linewidth]{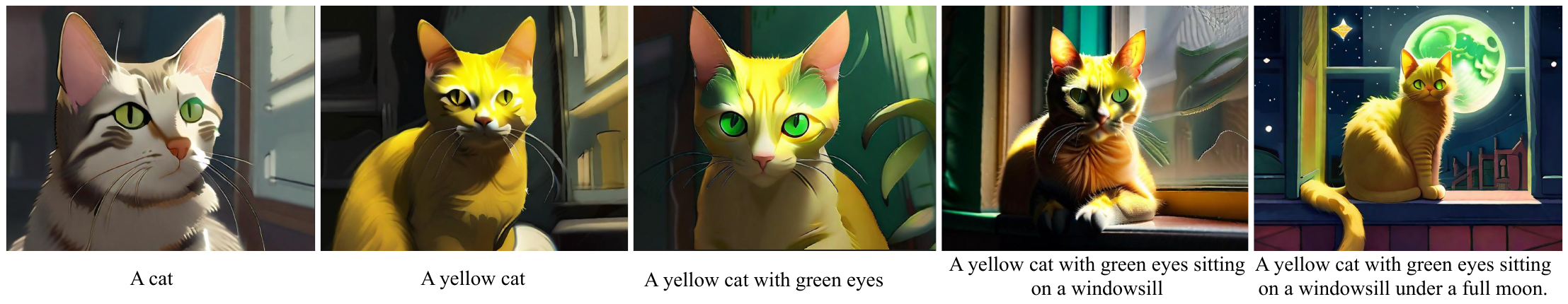}
    \end{center}
    \caption{\textbf{Visualization Comparison for Detailed Concepts.} ParaDiffusion exhibits superior alignment capabilities, effectively managing the complexity of multiple objects and attributes.}
    \label{fig:Concepts}
\end{figure*}

\begin{table}[t]
    \centering
    \small 
    \setlength{\tabcolsep}{1mm}
    \caption{\textbf{Ablation of Caption Length for ParaImage on ParaPrompts-400.} We prompt CogVLM~\cite{WeihanWang} to generate captions of various lengths. $500$k images of ParaImage-Big are randomly sampled from the original dataset. We only evaluate Text Faithfulness for the experiment. Longer captions improve significantly.
    }
    \scriptsize

    

\scriptsize
\begin{tabular}{l|c|cccc}

    \multirow{2}{*}{Dataset} & \multirow{2}{*}{ Size} & \multicolumn{4}{c}{ Average Length of Caption ($n$ words)}\\
\cline{3-6} 
    & & 10 words & 50 words & 100 words & 150 words \\
    \shline
    \hline
    ParaImage-B & 500k & 8.5\% & 21.1\% & 29.8\% & \textbf{40.6}\%\\
    ParaImage-S & 3k & 16.4\% & 23.3\% & 29.1\% & \textbf{31.2}\%

    \end{tabular}
    \label{ablation_3}
\end{table}

\begin{table}[t]
    \centering
    \small 
    \setlength{\tabcolsep}{1mm}
    \caption{\textbf{Ablation Study for LLM Adaptation on ParaPrompts-400.} We only evaluate Text Faithfulness for the experiment. 
    `P.' refers to `Parameters'.}
    \scriptsize
\begin{tabular}{l|c|ccc}
    Hyperparameters  & \# Trainable P. & Win &  Tie & Lose \\
    \shline
    \hline
    w/o LoRA~(Base)& - & -& -& -\\
    w LoRA~(\( r \)=8, \( \alpha \)=32) & 4.2M~(0.06\%)& 50.6\% & 4.3\% & 45.1\% \\
    w LoRA~(\( r \)=16, \( \alpha \)=32) & 8.4M~(0.13\%)& 50.8\% & 2.6\% & 46.6\% \\
    w LoRA~(\( r \)=32, \( \alpha \)=32) & 16.7M~(0.25\%)& 51.1\% & 3.6\% & 45.3\% \\
    w LoRA~(\( r \)=64, \( \alpha \)=32) & 33.6M~(0.51\%)& 53.4\% & 2.1\% & 44.5\% \\
    w LoRA~(\( r \)=128, \( \alpha \)=32) & 67.1M~(1.01\%) & 51.9\% &  2.8\% & 45.3\% \\
    \hline
    w LoRA~(\( r \)=64, \( \alpha \)=16) & 33.6M~(0.51\%)& 52.3\% & 2.5\% & 45.2\% \\
    w LoRA~(\( r \)=64, \( \alpha \)=32) & 33.6M~(0.51\%)& \textbf{53.4\%} & \textbf{2.1\%} & \textbf{44.5\%} \\
    w LoRA~(\( r \)=64, \( \alpha \)=64) & 33.6M~(0.51\%)& 53.1\% & 2.8\% & 44.1\% 
    \end{tabular}
    \label{ablation_1}

\end{table}



\begin{table}[t]
    \centering
    \small 
    \setlength{\tabcolsep}{1mm}
    \caption{\textbf{Definition and explanation of LoRA Hyperparameters \( r \) and \( \alpha \).}
    }
    \scriptsize
\begin{tabular}{p{2cm}|p{2.5cm}|p{3cm}}
   Params & Definition & Effect \\
    \shline
    \hline
    \( r \) (LoRA Rank) & Dimension of low-rank decomposition & Higher \( r \) improves expressivity but increases parameters and computation. \\
    \hline
    \( \alpha \) (LoRA Alpha) & Scaling factor for LoRA updates & Higher \( \alpha \) strengthens LoRA’s impact but may affect stability. \\
    \hline
    \( \frac{\alpha}{r} \) (Scaling) & Scaling applied to \( AB \) update & Adjusts the update magnitude, balancing adaptation and stability. \\
    \end{tabular}
    \label{tab:lora_hyperparams}

\end{table}

\begin{table}[t]
    \centering
    \small 
    \setlength{\tabcolsep}{1mm}
    \caption{\textbf{Ablation for Image Quality on ParaPrompts-400.} $3$k training data is used for all datasets. We only evaluate Visual Appeal.
    }
    \scriptsize

    

\scriptsize
\begin{tabular}{l|c}

    Dataset & Voting\\
    \shline
    \hline
    LAION &  3.1\%\\
    LAION-Aesthetics & 18.5\% \\ 
    ParaImage-Big & 23.7\%\\
    ParaImage-Small & \textbf{54.7}\% 

    \end{tabular}
    \label{ablation23}
\end{table}

\begin{table}[t]
    \centering
    \small 
    \setlength{\tabcolsep}{1mm}
    \caption{\textbf{Ablation for Data Size of ParaImage-Big on ParaPrompts-400~(Human Voting) and MS-COCO $256\times 256$~(FID-30K).}  We only evaluate Visual Appeal for Human Voting. We randomly sample the data to a smaller scale.
    }
    
\scriptsize
\begin{tabular}{l|c|c}

    Data Size (ParaImage-Big) & Human Voting & FID-30K$\downarrow$ \\
    \shline
    \hline
    0.3m & 17.1\% & 13.45 \\ 
    1.3m & 32.7\% &  10.23\\
    3.3m & \textbf{50.2}\%  & \textbf{9.64}

    \end{tabular}
    \label{ablation4}
\end{table}

\subsubsection{Layout-based Image Generative v.s. Text-based Image Generation}
The current approaches to paragraph-to-image generation can be broadly categorized into two main strategies. The first is layout-based image generation, which utilizes models like ChatGPT or GPT-4 to generate bounding box layouts as the condition for image generation, as illustrated in Figure~\ref{fig:comparison_layout}. 
The second approach is our pure text-to-image alignment method for paragraph-to-image generation. 

To better compare and analyze these two paradigms, we provide corresponding experimental results and visualizations, as shown in Figure~\ref{fig:layout} and Figure~\ref{fig:vis_layout}.
LLM Blueprint~\cite{gani2023llm}, the SOTA layout-based model, as a baseline for comparison with our proposed model.
Figure~\ref{fig:layout} demonstrates that our model significantly outperforms LLM Blueprint in both Text Faithfulness and Visual Appeal. 
The reasons for this improvement can be summarized as follows:
1) \textbf{Stronger foundation model}: Layout-based models, such as LLM Blueprint, rely on Stable Diffusion 1.5 as the foundation model, which is less effective in text alignment and visual appeal compared to our ParaDiffusion.
2) \textbf{Limitations of box-conditioned generation}: Box-conditioned layouts are not always optimal for complex text alignment.
As shown in Figure~\ref{fig:ablation}(c) and Figure~\ref{fig:comparison_layout}, box-in-box scenarios frequently arise, making precise alignment challenging. 
Specifically, we summarize the following drawbacks of layout-based models:
\textit{Overlap leads to misalignment in object count:} When two identical objects overlap excessively, it often results in generating only one object. For example, two buildings may be represented as just one due to excessive overlap.
\textit{Overlap causes object loss:} When one object is inside another, it becomes difficult to generate the inner object. For instance, in the case of 'people in the boat,' the person may fail to be generated.
\textit{Incorrect relationships between objects:} When two objects overlap, it can create issues in generating accurate object interactions. For example, 'a Samoyed lying on stones' may be incorrectly generated as 'a Samoyed standing on stones.
%
Therefore, current box-conditioned image generation models struggle to handle such overlapping relationships effectively.
Additionally, as shown in Figure~\ref{fig:comparison_layout}, layout-based models perform poorly in visual appeal. 
Many cases cannot be used, especially in LLM-grounded diffusion models, where a significant number of images are of poor quality.

\subsubsection{Evaluation for The Detailed Concepts}
In addition to exploring the overall assessment of Text Faithfulness, we also evaluated three more granular aspects: Objects, Attributes, and Locations, as shown in Figure~\ref{parts}.
$50$ samples were randomly selected from the ParaPrompts 400 dataset, and we prompted GPT-4 to extract the Objects, Attributes, and Locations from each image description, as shown in Figure~\ref{pipeline-shot}.
This allowed us to obtain the number, names, attributes, and locations of objects in each prompt description.
An evaluator was then tasked with providing a distribution score for each aspect, on a scale from $1$ to $5$, with $5$ being the highest.
The results reveal that ParaDiffusion outperforms LLM Blueprint~\cite{gani2023llm} in terms of alignment with Objects and Attributes, while it performs slightly worse in terms of Location alignment.
The primary reason for this outcome is similar to previous analyses: layout-based image generation models, which use box-conditioned image generation (Paint by Example~\cite{Yang_2023_CVPR}, BoxDiff~\cite{Li_2023_CVPR}), achieve precise alignment for locations. 
However, these models also have significant limitations, such as issues with "box-in-box" scenarios, which prevent them from effectively generating and fitting all objects and attributes.
The corresponding object, attribute, and location annotations can be found at \href{https://github.com/weijiawu/ParaDiffusion/tree/main/ParaPrompts-400/Parts_50}{\color{blue}{\tt ParaPrompts-50-Concepts}}.

Additionally, to delve deeper into the comprehension and alignment of nuanced concepts of ParaDiffusion, along with the precision of object and attribute alignment, Figure~\ref{fig:Concepts} illustrates that our model adeptly tracks the semantic progression of the prompt.
By sequentially incorporating pertinent object and attribute elements, ParaDiffusion demonstrates its ability to efficiently align with the intricacies of the textual instructions provided.

\begin{table*}[t]
    \centering
    \small 
    \setlength{\tabcolsep}{1mm}
    \caption{\textbf{Performance Comparison at Different Stages on ParaPrompts-400.} 
    }
    \scriptsize
\begin{tabular}{l|cc|cc}

    \multirow{2}{*}{Dataset}  &  \multicolumn{2}{c|}{Visual Appeal} &  \multicolumn{2}{c}{Text Faithfulness} \\
    & Win (\%)  & Lose (\%)  & Win (\%)  & Lose (\%) \\
    \shline
    \hline
    Pre-Train, Stage-1 &  - &  - & - & - \\
    Stage-2 $vs.$ Stage-1~(w ParaImage-B)  & 82.1 & 17.9 & 76.9 & 32.1 \\
    Stage-3 $vs.$ Stage-2~(w ParaImage-S)   & 86.2 &13.8 & 53.1 & 46.9\\
    \end{tabular}
    \label{ablation_2}
\end{table*}

\begin{table*}[t]
    \centering
    \small 
    \setlength{\tabcolsep}{1mm}
    \caption{\textbf{Evaluation for Generated Caption.} We used different model to generate captions for 500 images, scoring them from three perspectives on a scale of $1$ to $5$, with $5$ being the best. 
    }
    \scriptsize
\begin{tabular}{l|ccc|c}
    Model  & Description Completeness & Relevance &  Fantasy & Ave. Score \\
    \hline
    CogVLM~\cite{WeihanWang} & 4.454 & 4.297 & 4.162 & 4.304\\
    LLaVA 1.5~\cite{liu2024improved} & 3.902 & 4.236 & 4.212 & 4.117 \\
    MiniGPT-4~\cite{zhu2023minigpt} & 3.744 & 3.523 & 3.233 & 3.500 \\
    \end{tabular}
    \label{GeneratedCaption}
\end{table*}

\subsection{Ablation Study}

\subsubsection{Effect of Language Model Adaptation.}
Table~\ref{ablation_1} presents the ablation study for LLM Adaptation.
`Base' denotes directly performing paragraph-image alignment learning~(Stage-2) with ParaImage-Big without fine-tuning Llama V2 using LoRA.
Compared to the base model, our LLM adaptation with LoRA  demonstrates an improvement of nearly $5\%$ in human voting rates.
Another insight reveals that during the process of increasing trainable parameters from $4.2$ million to $67.1$ million, the performance appears to gain consistently. 
Therefore, in our final configuration, we randomly select 16.7 million trainable parameters as the ultimate setting.
Figure~\ref{fig:ablation} (b) presents the visual comparisons for the performance from LLM adaptation.
\( r \) and \( \alpha \) are two key hyperparameters, representing LoRA Rank and LoRA Alpha, respectively. Table~\ref{tab:lora_hyperparams} provides their detailed definitions and explanations. The performance comparison for different hyperparameter settings is presented in Table~\ref{ablation_1}, where the optimal configuration is identified. The results indicate that when \( r \)=64, \( \alpha \)=32, the model achieves the best performance.

\subsubsection{Effect of Caption Length.}
Table~\ref{ablation_3} presents the ablation study of caption length for the proposed dataset, ParaImage.
We control the same other variables, such as the size of training data and the images, and only change the corresponding caption length~(words) through prompt CogVLM~\cite{WeihanWang}. 
The specific prompt is as follows: `Provide a detailed description of the image using approximately $n$ words.'
It can be observed that as the length of the caption increases, the performance of the generated images in terms of text faithfulness also gradually increases.
The data with average captions of around 150 words achieved the highest proportions of human voting, with $40.6\%$ and $31.2\%$ for ParaImage-Big and ParaImage-Small, respectively.
It appears that an average of 150 words per caption still has not reached the peak performance. 
However, captioning capability of CogVLM is limited, unable to capture more granular descriptive information (exceeding average 150 words for the whole dataset).

\subsubsection{Effect of Image Quality.}
Table~\ref{ablation23} presents the ablation for image quality of ParaImage.
To ensure a fair comparison, for all datasets, we randomly sample $3$k data from the original dataset as the training set for training.
It can be observed that the proposed ParaImage-Small exhibits the best performance, the models trained on this data achieving the highest human voting for visual appeal, at $54.7\%$.
We believe this is reasonable, as the 3k images of ParaImage-Small were manually selected by annotators from a pool of 650k high-quality images of LAION-Aesthetics.

\subsubsection{Effect of Data Size.} Table~\ref{ablation4} presents the ablation study on the impact of data size. As the data size increases, performance of the model improves significantly across both automated metrics and human voting evaluations.
Specifically, the model trained on 3.3 million ParaImage-Big data surpasses the one trained on 1.3 million, achieving a 17.5\% improvement in human voting. 
Similarly, in the automated metric FID-30K, the score improves notably from 13.45 to 9.64.

\subsubsection{Performance Comparison for Different Stages.}
Table~\ref{ablation_2} presents performance comparison at different stages.
ParaImage-Big brings significant improvements, with approximately a $50\%$ increase in human voting rates for both visual appeal and text faithfulness simultaneously.
The smaller-scale, and high-quality set of 3k images in ParaImage-Small further enhances aesthetic aspects, achieving a remarkable around $70\%$ increase in human preference rates compared to that of Stage 2.
Figure~\ref{fig:ablation} (a) presents a visual comparison, vividly illustrating the gains in visual appeal. 
In this case, the prompt is `An exquisite sculpture in the center of a square on a rainy day.'


\subsubsection{Quality Evaluation for Generative Captions of ParaImage-Big}
As demonstrated in Section~\ref{generativecaption}, we utilized a substantial amount of synthetic caption data.
However, the quality of synthetic captions still leaves room for improvement and does not match the precision of manually annotated captions (ParaImage-Small).
Here, we assessed the quality of synthetic captions. As indicated in Table~\ref{GeneratedCaption}, we primarily selected three mainstream vision-language models (CogVLM, LLaVA, MiniGPT-4) to generate corresponding captions, which were then manually evaluated for quality.
We mainly evaluated the attributes of the captions (Description Completeness, Relevance, Fantasy) from three aspects. 
In this evaluation, three assessors were invited.
For each generated caption, they were scored on a scale from 1 to 5, and we took the average of their scores as the final score for that caption.
Among these three models, CogVLM performed the best, achieving a score of $4.304$.
\begin{figure*}[tbp]
    \begin{center}
        \includegraphics[width=.98\linewidth]{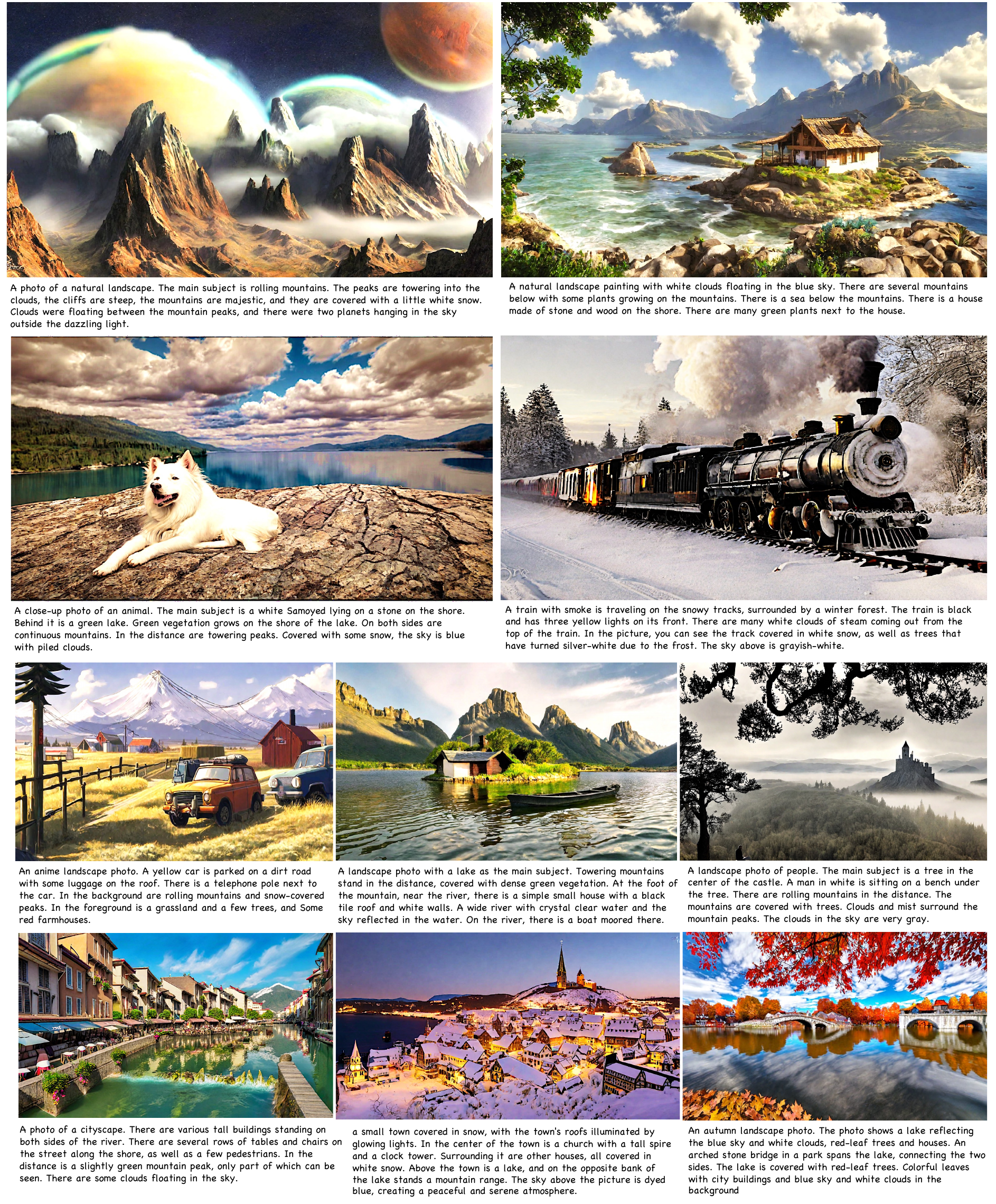}
    \end{center}
    \vspace{-0.7cm}
    \caption{\textbf{More Visualizations for \textcolor{red}{scenery-centric} from ParaDiffusion.}}
    \label{fig:fig2}
    \vspace{-0.2cm}
\end{figure*}

\begin{figure*}[tbp]
    \begin{center}
        \includegraphics[width=.98\linewidth]{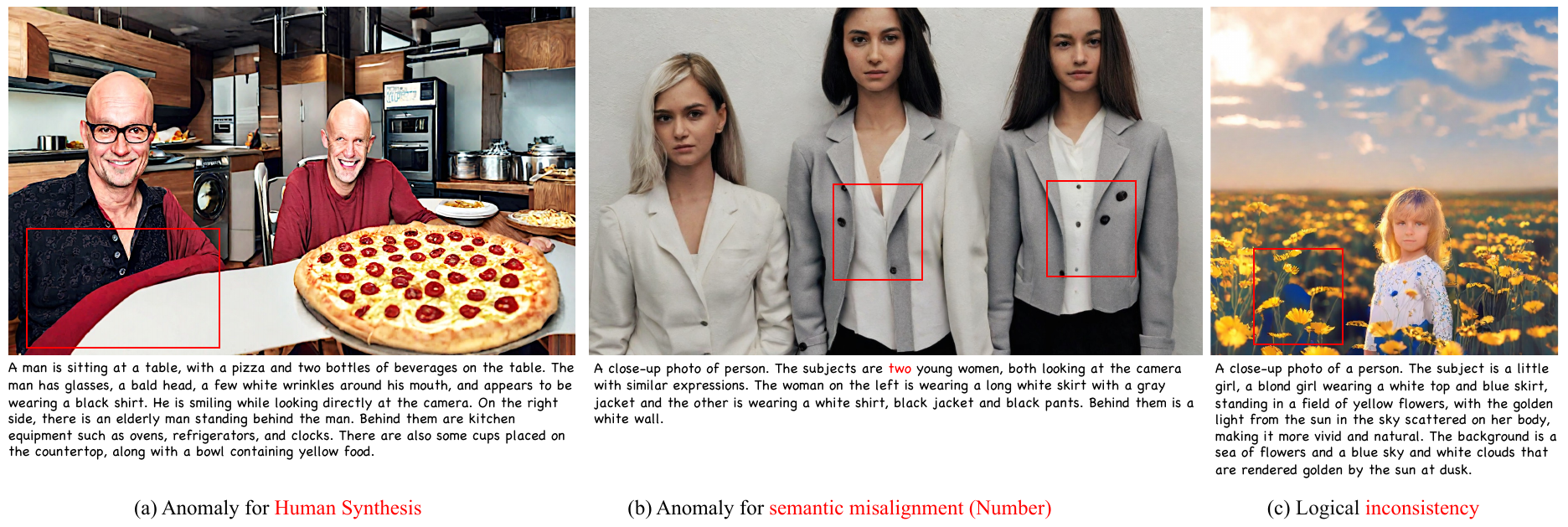}
    \end{center}
    \vspace{-0.5cm}
    \caption{\textbf{Bad Cases for ParaDiffusion.} 
    There are still some areas that can be optimized for our ParaDiffusion.}
    \label{fig:badcase}
    \vspace{-0.2cm}
\end{figure*}

\begin{figure}[tbp]
    \begin{center}
        \includegraphics[width=.98\linewidth]{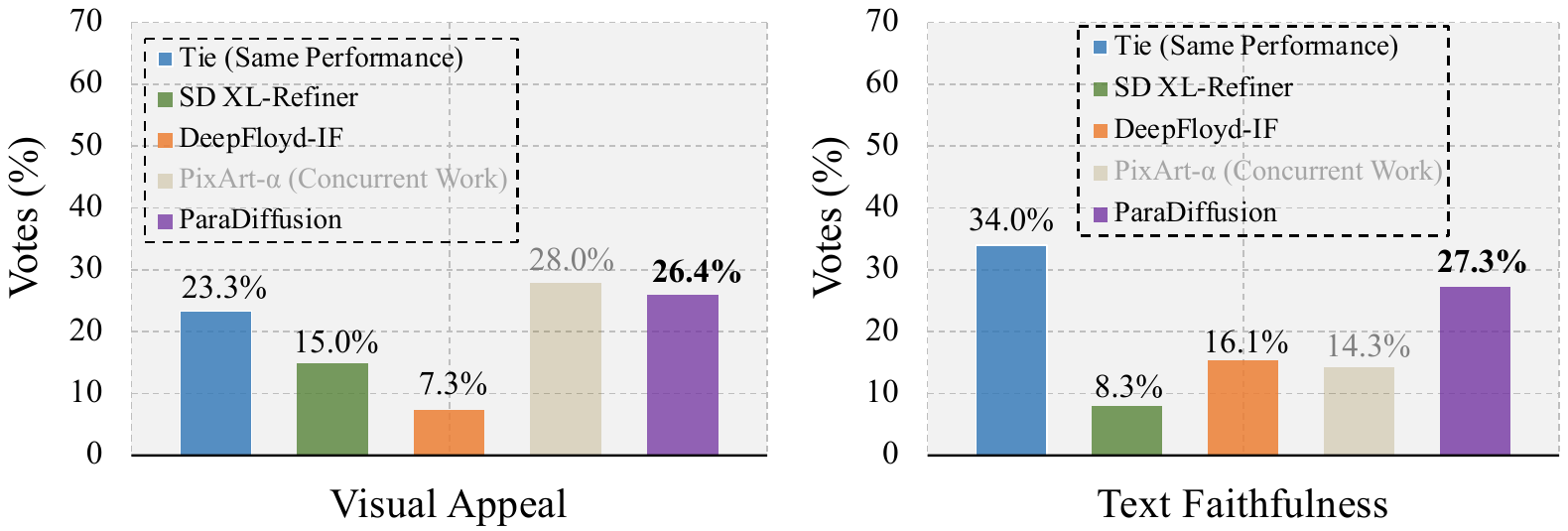}
    \end{center}
    \vspace{-0.5cm}
    \caption{\textbf{User study on 1600 prompts of PartiPrompts.} We only selected open-source models for comparison, when the results of closed-source models~\cite{xue2023raphael,radford2021learning} were unavailable or API calls failed.}
    \label{fig:PartiPrompts}
    \vspace{-0.2cm}
\end{figure}

\subsubsection{Potential Biases in Human Evaluation}
Human evaluation is crucial for assessing generative models, but it is important to acknowledge several potential biases that may influence the results:
\begin{itemize}
    \item \textbf{Evaluator Bias}. Different evaluators may have varying personal preferences, cultural backgrounds, or levels of expertise, which could affect their ratings. To reduce this bias, one effective approach is to ensure diversity in our evaluator pool and provide clear, standardized rating criteria.
    \item \textbf{Rule Bias}. Evaluators may interpret the rating scale differently, leading to inconsistencies in scores. To address this, one effective approach is to provide detailed guidelines and examples for each score level, along with a calibration process to align evaluators’ understanding of the scale. For example, in Sections~\ref{SelectionRule} and~\ref{manualcaptions}, we implemented constraints on caption labeling and image selection to enhance objectivity and reduce individual subjectivity.
    \item \textbf{Task Familiarity Bias}: Familiarity with the task can affect how evaluators assess the generated images. Experts might focus on technical details, while novices may prioritize overall aesthetics. 
\end{itemize}
While human evaluations are essential, they are subject to various biases.
Therefore, in conducting our user study, we carefully considered these biases and established comprehensive evaluation rules to minimize subjective influences. 
It is also important to expand the evaluator pool to ensure diversity across different backgrounds, aiming to make the evaluation results more objective and fair.

\section{Research Data Policy and Data Availability Statements}
\begin{itemize}
\itemsep -0cm
    \item 
    The datasets used during and/or analysed during the current study are available in the related public dataset repositories, according to Sec.~\ref{Dataset}. Or more detailed information can be found at our repository \href{https://github.com/weijiawu/ParaDiffusion}{\color{blue}{\tt https://github.com/weijiawu/ParaDiffusion}}.
    
   \item  
   For the datasets introduced in this paper (\ie ParaImage), we have open-sourced the training and testing sets, as well as the corresponding methods for data collection, screening, annotation, and evaluation. Please refer to the respective dataset GitHub link for specific details.
   
\end{itemize}

\section{Conclusion}
\label{sec:conclusion}
In this paper, we firstly present the challenges of long-text alignment in the task of text-guided image generation. 
Additionally, we propose an effective solution that addresses both the data and algorithmic aspects. 
In terms of algorithms, we introduce an information-enriched diffusion model, which explores the transfer of the long-text semantic understanding capabilities from large language models to the image generation task.
For data, we construct and present a high-quality, textual rich paragraph-to-image pairs dataset, where the corresponding textual descriptions can extend up to $400$ words.
The experiments demonstrate that our ParaDiffusion achieves outstanding performance in both visual appeal and text faithfulness aspects.

\section{Appendix}

\subsection{More Visualizations}
\begin{figure*}[tbp]
    \begin{center}
        \includegraphics[width=.98\linewidth]{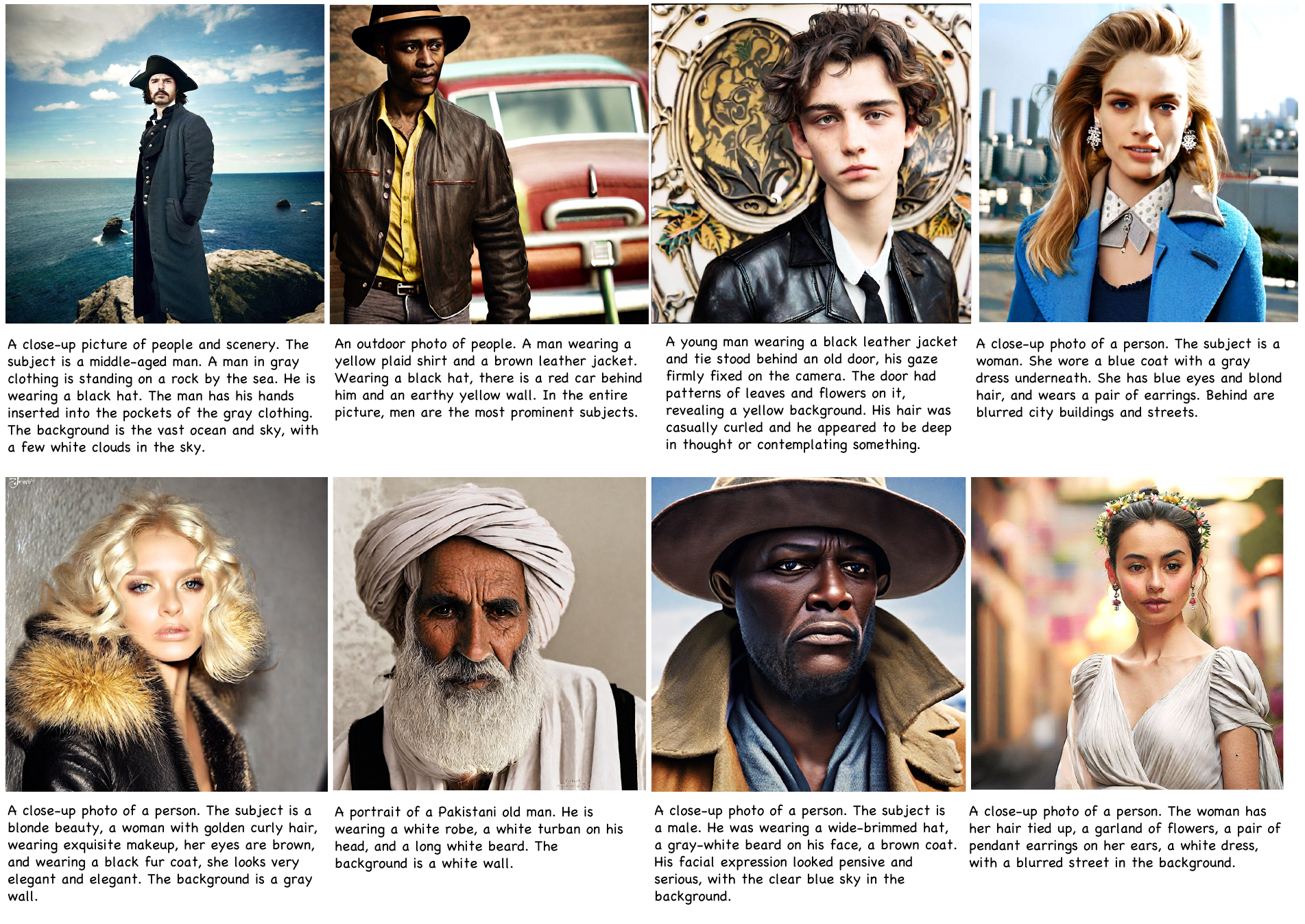}
    \end{center}
    \vspace{-0.5cm}
    \caption{\textbf{More Visualizations for human-centric from ParaDiffusion.}}
    \label{fig3.pdf}
    \vspace{-0.2cm}
\end{figure*}

Figure~\ref{fig3.pdf} and Figure~\ref{fig:fig2} provides more visualizations of ParaDiffusion for human-centric and scenery-centric domains, respectively.
The visualization reveals that our ParaDiffusion can generate intricate and realistic composite images of individuals as well as picturesque scenes.
Moreover, it is noteworthy that images generated through paragraph-image generation exhibit a compelling narrative quality, enriched with profound semantics.
The generated images consistently feature intricate object details and demonstrate effective multi-object control.

\subsection{Limitations}
Despite ParaDiffusion achieving excellent performance in long-text alignment and Visual Appeal, there are still some areas for improvement, such as inference speed.
ParaDiffusion has not been optimized for speed, and implementing effective strategies, such as ODE solvers~\cite{lu2022dpm} or consistency models~\cite{song2023consistency}, could lead to further enhancement and optimization in inference speed.
In addition, while ParaDiffusion exhibits the capability to produce images of high realism, the presence of undesirable instances persists, as depicted in Figure~\ref{fig:badcase}.
Two predominant strategies prove effective in addressing these challenges: 
Firstly, at the data level, augmenting the dataset with additional high-quality images enhances diversity, contributing to further model refinement. 
Secondly, at the algorithmic level, the incorporation of additional constraints, such as geometric and semantic constraints, serves to imbue synthesized images with greater logical and semantic coherence.

\begin{figure*}[tbp]
    \begin{center}
        \includegraphics[width=.98\linewidth]{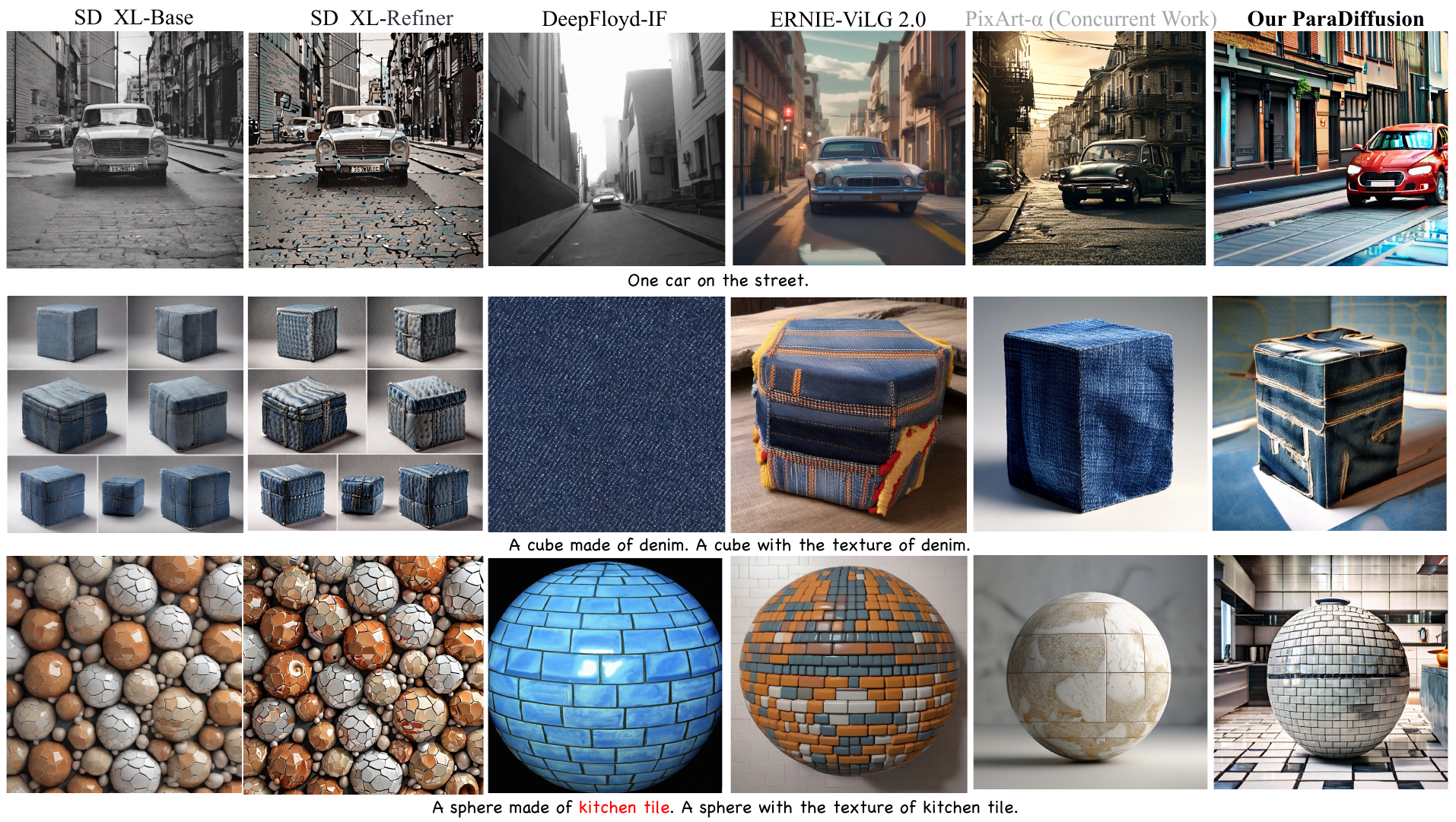}
    \end{center}
    \vspace{-0.5cm}
    \caption{\textbf{Visualization Comparison on ViLG-300.} Our ParaDiffusion exhibits competitive performance in visual appeal.}
    \label{fig:all_ViLG}
\end{figure*}

\begin{figure*}[tbp]
    \begin{center}
        \includegraphics[width=.98\linewidth]{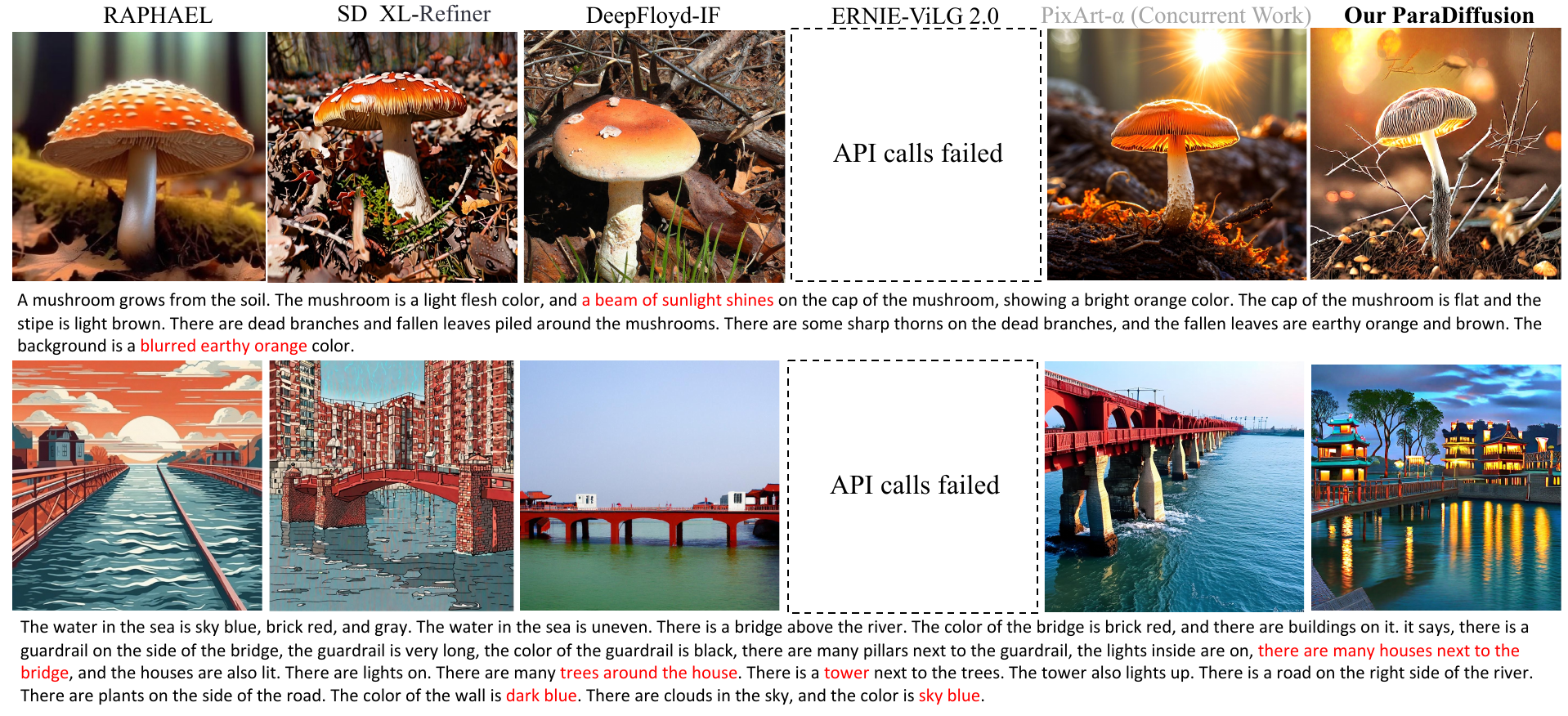}
    \end{center}
    \caption{\textbf{Visualization Comparison on ParaPrompts-400.} Our ParaDiffusion demonstrates significant advantages in long-text alignment.}
    \label{fig:ParaPrompts_vis}
\end{figure*}

\begin{figure*}[tbp]
    \begin{center}
        \includegraphics[width=.98\linewidth]{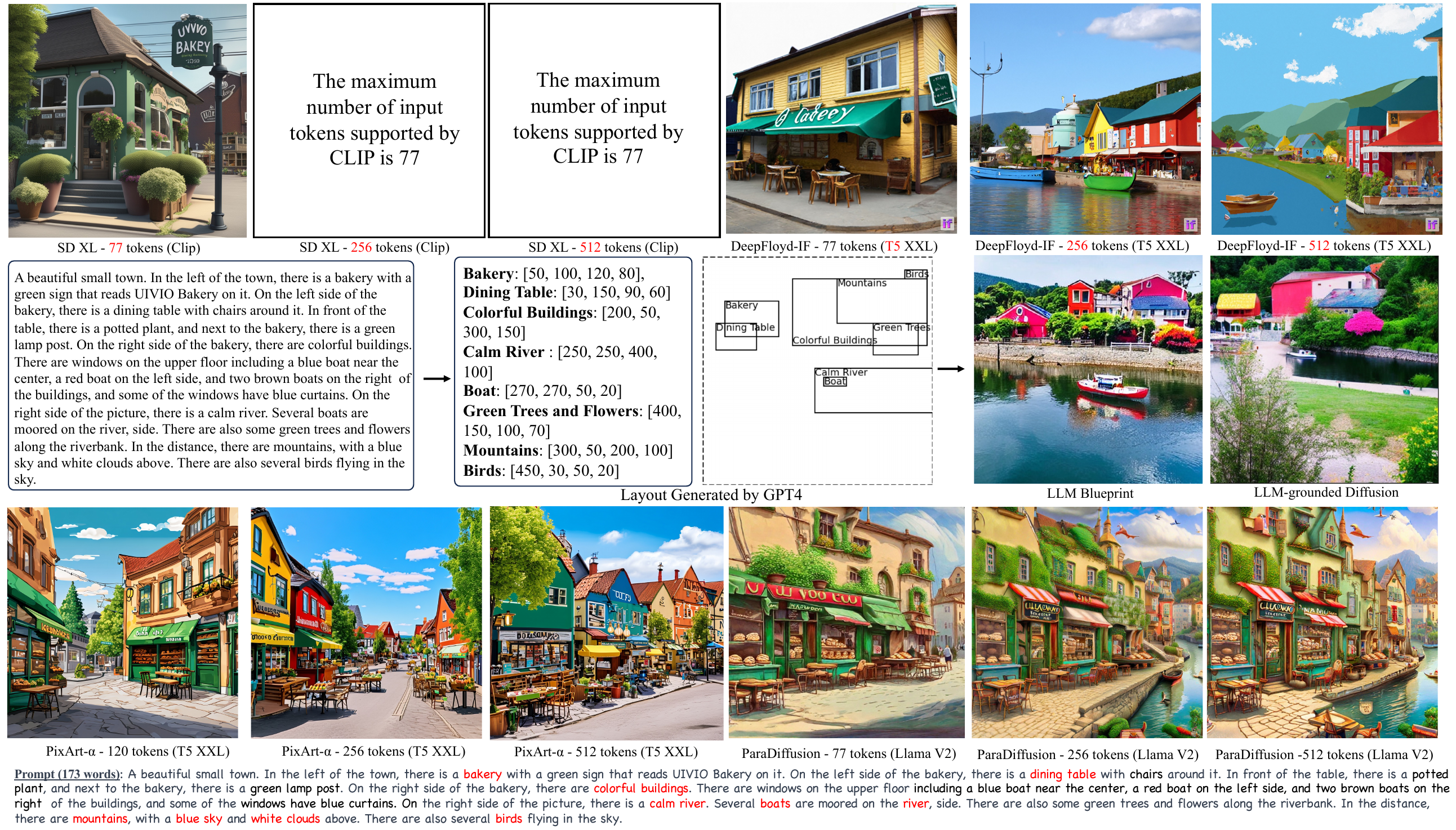}
    \end{center}
    \vspace{-0.5cm}
    \caption{\textbf{Risk of Conflict between Visual Appeal and Text Faithfulness.} Two insights:1) Merely extending the token count of the current model~(SD XL, DeepFloyd-IF)) does not yield satisfactory performance. 2) As the number of input tokens increases, all models experience a certain degree of decline in visual appeal.}
    \label{fig:DeepFloyd}
    \vspace{-0.2cm}
\end{figure*}

\subsection{Experiments on 1,600 prompts of PartiPrompts}
We also provides the related experiment results on \\PartiPrompts-1600, as shown in Figure~\ref{fig:PartiPrompts}.
It can be observed that our model also achieved outstanding performance in Text Faithfulness, with a $27.3\%$ human voting rate, significantly outperforming previous models such as SD XL and DeepFloyd-IF. 
Additionally, our model demonstrated a competitive advantage in visual appeal, surpassing SD XL and DeepFloyd-IF, and approaching the performance of contemporaneous work PIXART-$\alpha$.
A notable observation in ParaPrompts is the high proportion of 'Tie' votes in human evaluations, especially for Text Faithfulness, with a voting rate of up to $34\%$.
This is attributed to the presence of numerous simple or abstract prompts in ParaPrompts-1600, making it challenging to provide precise voting results, such as for prompts like `happiness' or `emotion.'

\subsection{Visualization Comparison on ViLG-300 and ParaPrompts-400}
To offer a more intuitive comparison, we provide visualizations comparing our model with prior works on ViLG-300 and ParaPrompts-400 datasets, as depicted in Figure~\ref{fig:all_ViLG} and Figure~\ref{fig:ParaPrompts_vis}.
From the perspective of visual appeal, the synthesized images produced by our ParaDiffusion align well with human aesthetics.
They exhibit qualities reminiscent of photographic images in terms of lighting, contrast, scenes, and photographic composition.
Concerning the alignment of long-form text, our ParaDiffusion demonstrates outstanding advantages, as illustrated in Figure~\ref{fig:ParaPrompts_vis}.
Previous works often struggle to precisely align each object and attribute in lengthy textual descriptions, as seen in the second row of Figure~\ref{fig:ParaPrompts_vis}. 
Existing models frequently miss generating elements like `towers' and `houses', and their relative spatial relationships are flawed.
In contrast, our model excels in accurately aligning the textual description with the content in the image.

\subsection{More Details for ParaImage-Small}
\label{SelectionRule}

As stated in the main text, we selected $3,000$ exquisite images from a pool of $650,000$ images curated by LAION-Aesthetics \cite{laion_aesthetics}, adhering to common photographic principles.
The detailed \textbf{Aesthetic Image Selection Rule} is outlined as follows:
\begin{itemize}
    \item The selected images will be used to annotate long-form descriptions (128-512 words, 4-10 sentences). Please assess whether the chosen images contain sufficient information (number and attributes of objects, image style) to support such lengthy textual descriptions.
    \item The images should not include trademarks, any text added in post-production, and should be free of any mosaic effects.
    \item \textbf{Spatial Relationships between Multiple Objects}: For images with multiple objects, there should be sufficient spatial hierarchy or positional relationships between these objects. For example, in a landscape photograph, the spatial distribution of mountains, lakes, and trees should create an interesting composition. There should be clear left-right relationships between multiple people.
    \item \textbf{Interaction between Multiple Objects}: For images with multiple objects, choose scenes that showcase the interaction between the objects. This can include dialogue between characters, interactions between animals, or other interesting associations between objects.
    \item \textbf{Attribute of Single Object}: All key details of the main subject should be clearly visible, and the subject's attribute information should include at least three distinct aspects, including color, shape, and size. For example, in wildlife photography, the feather color, morphology, and size of an animal should be clearly visible.
    \item \textbf{Colors, Shapes, and Sizes of Objects}: Various objects in the image should showcase diversity in colors, shapes, and sizes. This contributes to creating visually engaging scenes.
    \item \textbf{Clarity of the Story}: The selected images should clearly convey a story or emotion. Annotators should pay attention to whether the image presents a clear and engaging narrative. For example, a couple walking on the street, a family portrait, or animals engaging in a conflict.
    \item \textbf{Variety of Object Categories}: A diverse set of object categories enriches the content of the image. Ensure that the image encompasses various object categories to showcase diversity. For instance, on a city street, include people, cyclists, buses, and unique architectural structures simultaneously.
\end{itemize}

Following the aforementioned rules, we instructed the annotators to rate the images on a scale of 1-5, with 5 being the highest score.
Subsequently, we selected images with a score of 5 as the data source for ParaImage-Small, resulting in approximately 3k images.

\subsection{Risk of Conflict between Visual Appeal and Text Faithfulness}
We also explored the potential of existing architectures~(\textit{e.g.,} SD XL, DeepFloyd-IF) for long-text alignment of text-image image generation, as shown in Figure~\ref{fig:DeepFloyd}.
Firstly, all methods that utilize CLIP as a text encoder, such as SDXL, face limitations in supporting paragraph-image tasks due to the maximum number of input tokens supported by CLIP being $77$.
Secondly, we investigated the performance of methods using T5 XXL as a text encoder, \textit{e.g.,} DeepFloyd-IF~\cite{DeepFloyd} and PIXART-$\alpha$~\cite{chen2023pixart}.
We directly adjusted the tokens limitation of these methods to accommodate longer lengths, enabling support for image generation settings involving the alignment of long-form text.
With an increase in the number of tokens, the visual appeal quality of DeepFloyd-IF experiences a significant decline, becoming more cartoonish around $512$ tokens. 
Furthermore, its semantic alignment is unsatisfactory, with many generated objects missing, such as the table.
Similarly, PIXART-$\alpha$ fails to achieve satisfactory semantic alignment even with the maximum token limit increase, and its visual appeal also experiences a certain degree of decline.
In contrast, our ParaDiffusion exhibits a more stable performance, achieving good semantic alignment with $256$ tokens and showing minimal decline in visual appeal as the token count increases.
{
    \small
    \bibliographystyle{ieeenat_fullname}
    \bibliography{paper}
}

\end{document}